%% file: main.tex
\documentclass{article}
\PassOptionsToPackage{numbers, compress}{natbib}
\usepackage[preprint]{neurips_2026}
\usepackage{wrapfig}
\usepackage{multirow}
\usepackage{tabularx}
\usepackage{bbding}
\usepackage{caption}
\usepackage[ruled,vlined,linesnumbered]{algorithm2e}
\SetAlCapSkip{0.3em}
\SetAlgoSkip{0.4em}
\SetInd{0.3em}{0.8em}
\SetNlSkip{0.25em}
\SetAlgoNlRelativeSize{-1}

\DontPrintSemicolon
\SetKwInput{KwInput}{Input}
\SetKwInput{KwOutput}{Output}
\SetKwComment{Comment}{$//$ }{}
\SetAlgoSkip{small}
\SetAlgoInsideSkip{small}
\DontPrintSemicolon

\usepackage[utf8]{inputenc} 
\usepackage[T1]{fontenc}    
\usepackage{hyperref}       
\usepackage{url}            
\usepackage{booktabs}       
\usepackage{amsfonts}       
\usepackage{nicefrac}       
\usepackage{microtype}      
\usepackage{tabularx}
\usepackage{siunitx}
\usepackage{amsmath}
\usepackage{mathrsfs}
\usepackage[ruled,vlined,linesnumbered]{algorithm2e}
\usepackage[table]{xcolor}  
\usepackage{adjustbox}
\usepackage{pifont}
\usepackage{enumitem}
\usepackage{capt-of}
\usepackage{makecell}

\title{Physics-Informed Video Generation via Mixture-of-Experts Latent Alignment}

\author{%
  \normalfont
  {\bfseries
  Cong Wang\textsuperscript{1,2,3 *},\quad
  Hanxin Zhu\textsuperscript{4 *},\quad
  Jiayi Luo\textsuperscript{3,5},\quad
  Yonglin Tian\textsuperscript{1}}\\
  {\bfseries
  Xiaoqian Cheng\textsuperscript{3,4},\quad
  Peiyan Tu\textsuperscript{3,6},\quad
  Xin Jin\textsuperscript{3,7},\quad
  Long Chen\textsuperscript{1 $\dagger$},\quad
  Zhibo Chen\textsuperscript{3,4 $\dagger$}}\\ [0.6em]
  \textsuperscript{1}CASIA\quad
  \textsuperscript{2}UCAS\quad
  \textsuperscript{3}ZGCA\quad 
  \textsuperscript{4}USTC\quad 
  \textsuperscript{5}BUAA\quad 
  \textsuperscript{6}ZJU\quad 
  \textsuperscript{7}EIT \\
  \textsuperscript{*}Equal contribution 
  \textsuperscript{$\dagger$}Corresponding author
}

\begin{document}

\maketitle

\input{sec/0_abstract}   
\input{sec/1_intro}
\input{sec/2_related}

\input{sec/3_method}

\input{sec/4_exp}

\input{sec/5_conclu}


\clearpage

{\small
\bibliography{main}
}
\input{sec/X_suppl}

\end{document}

%% file: sec/0_abstract.tex
\begin{abstract}
    Large-scale video generation models have made remarkable progress in semantic consistency and visual quality, producing videos that are increasingly coherent and visually convincing. Nevertheless, the dynamics induced by pixel-level fitting do not naturally accommodate the regularities that govern real-world motion and interaction, resulting in persistent shortcomings in physical plausibility. To address this limitation, we propose \textbf{PILA} (Physics-Informed Latent Alignment), a framework that injects physics-structured latent guidance into the frozen flow-matching dynamics of pretrained video models. Specifically, PILA first employs anchored field estimation to map frozen-generator latents into an operational physical attribute bank organized by field-proxy slots, using observable motion as a kinematic anchor for constructing less directly observed proxies. To handle the heterogeneity of real-world dynamics, PILA adopts a mixture-of-experts design over physical categories. Label-prior masked expert routing selects category-specific operator experts, whose refinements are regularized by operational residuals abstracted from physical relations. Finally, the refined proxies are fused into the physical attribute bank and decoded into a correction to the flow-matching vector field, injecting physics-aware guidance while preserving the visual prior of the pretrained backbone. With staged adapter training on Wan 2.1-1.3B and direct transfer of the learned adapter to Wan 2.2-14B, PILA achieves state-of-the-art results on VBench-2.0, VideoPhy-2, and PhyGenBench in both visual quality and benchmark-measured physical plausibility. 
\end{abstract}

%% file: sec/1_intro.tex
\section{Introduction}
\label{sec:intro}

Recent large-scale video generation models have made impressive progress in visual fidelity, enabling the generation of high-quality videos from text prompts~\cite{ho2022video, blattmann2023stable, kong2024hunyuanvideo, yangcogvideox}. Nevertheless, these models still lack reliable mechanisms for representing physical regularities, often yielding collisions with implausible momentum transfer, gravity-inconsistent trajectories, temporally inconsistent material deformations, and non-causal environmental interactions. Such physical implausibility limits the utility of video generation models in applications that require faithful dynamics, such as embodied-agent training and physical process visualization.

Explicitly integrating physics simulators into generation pipelines~\cite{xie2024physgaussian, chen2025physgen3d, zhang2024physdreamer} offers a natural remedy, but also introduces important limitations. These approaches typically require accurate object level 3D geometry~\cite{li2025wonderplay} and carefully specified physical states, which are costly to obtain and fragile in unconstrained real world settings. Moreover, coupling simulators with high capacity generators adds substantial engineering complexity and remains challenging for realistic foreground and background interactions. This motivates learning physics aware representations directly from video data without explicit simulation.
Beyond simulator-based approaches, a growing body of work seeks to improve physical realism through implicit inductive biases in data-driven video generation. Existing methods have explored external physical reasoning and planning to guide synthesis toward more plausible dynamics~\cite{xue2025phyt2v,zhao2026phyrpr,yang2025vlipp}, representation-level or inference-time alignment to encourage physical consistency~\cite{yuan2026inference,shen2026phantom}, and mixture-of-experts designs to inject structured physical priors into the generation process~\cite{wang2025wisa,wang2025prophy}. Despite their differences, these approaches still introduce physics mainly through prompts, rewards, alignment signals, or coarse expert priors. The unresolved problem is not merely how to add physical supervision, but how to create an operational physical interface, where field-like latent proxies can be constructed, regularized, and propagated back to the latent vector field.

\begin{figure*}[t]
    \centering
    \includegraphics[width=1\linewidth]{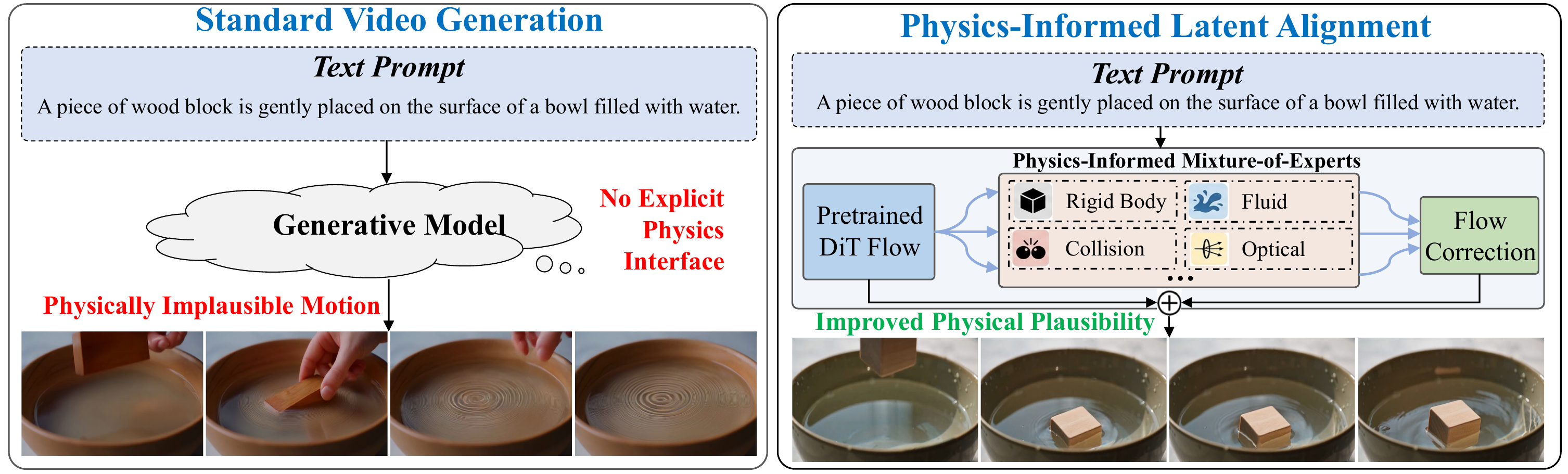}
    \caption{\textbf{Standard video generation vs. PILA}. Standard generators may produce physically implausible dynamics. PILA introduces a physics-structured latent adapter with routed experts and flow correction to improve physical plausibility.}
    \vspace{-0.96cm}
    \label{fig:teaser}
\end{figure*}

Motivated by this gap, we treat physics-aware generation as proxy-based latent alignment through an operational physical interface inside frozen flow-matching dynamics. Such an interface must construct field-like proxies from generator latents, refine them with category-specific residual structure, and translate the refined physical attribute bank back into the generator's vector field. The objective is not to reconstruct exact simulator states or calibrated physical measurements, but to expose a structured intermediate space where physics-derived constraints can shape a lightweight correction to the pretrained generator.

Physics-Informed Latent Alignment (PILA) instantiates this shared physical interface with an operational physical attribute bank. At each denoising step, a frozen-backbone latent estimate is encoded into fixed field-proxy slots for motion, pressure/density, thermal and phase support, strain/stress, impulse, and wave-like activity. An anchored field estimation (AFE) strategy uses observable motion as a kinematic anchor for completing less directly observed proxies. Label-prior masked expert routing (LPMER) then uses an LLM-assisted label parser to infer prompt-derived physical-category labels and select the corresponding operator experts. Each selected operator applies a recipe-masked residual update to the shared bank, and category-specific residual constraints (CSRC) regularize the corresponding field-proxy representations with operational residuals abstracted from physical relations. These residuals include PDE-style anchors, kinematic consistency terms, latent closure proxies, and stabilizing priors; they are not claimed to be complete governing equations on calibrated physical states. Finally, the refined physical attribute bank is decoded together with the frozen flow-matching prediction into a correction to the flow-matching vector field, injecting physics-aware guidance without replacing the pretrained visual backbone.
Our main contributions can be summarized as:
\begin{itemize}[leftmargin=*,labelsep=0.5em,itemsep=0.2em,topsep=0.2em]
    \item We propose a multi-expert physical latent-space constraint alignment framework for video generation that can be directly integrated into existing flow-matching generators, injecting physics-aware guidance through a lightweight latent correction.
    \item We introduce anchored field estimation to construct an operational physical attribute bank from generator latents, and design label-prior masked expert routing to activate the physical-category experts indicated by prompt-derived label priors.
    \item We propose field-proxy constraint alignment, where each routed expert reads the corresponding field-proxy representation from the attribute bank and is regularized by category-specific operational residuals.
    \item We demonstrate that the refined physical attribute bank provides an effective physics-aware correction signal for flow matching, yielding state-of-the-art motion continuity and benchmark-measured physical plausibility.
\end{itemize}

%% file: sec/2_related.tex
\section{Related Works}
\label{sec:relat}

\subsection{Data-driven general purpose video generation}

Data-driven general-purpose video generation learns open-domain video synthesis from large-scale video-text and image-video data, and has achieved substantial progress in visual fidelity, semantic alignment, and temporal coherence~\cite{ho2022video, blattmann2023stable, kong2024hunyuanvideo, yangcogvideox, wan2025wan}. Modern approaches are typically formulated within latent diffusion~\cite{blattmann2023stable, podell2023sdxl} or flow-matching~\cite{jin2024pyramidal} frameworks, which support scalable optimization and high-resolution video synthesis, and have increasingly emerged as the backbone of downstream adaptation~\cite{liang2024flowvid, wang2024worlddreamer, huang2025voyager} and controllable generation~\cite{hu2022make, wen2024panacea, shi2024motion}. Their strong performance is supported by large and diverse training data~\cite{nan2024openvid, geiger2012we, sun2020scalability, tan2024vidgen}, which expose the models to broad scene distributions, motion patterns, and semantic concepts in real-world videos. However, these models are primarily optimized for appearance quality and semantic consistency, with physical regularities learned only implicitly from data. As a result, they often produce visually convincing videos that still exhibit implausible motion, object interactions, and material dynamics.

\subsection{Explicit physics constrained video generation}

Explicit physics simulation methods reconstruct 3D scene representations and simulate dynamics using Material Point Methods or Position-Based Dynamics~\cite{xie2024physgaussian, chen2025physgen3d, liu2024physgen, lin2025omniphysgs, le2025gravity}. Recent extensions have broadened this paradigm along several directions. WonderPlay~\cite{li2025wonderplay} and RealWonder~\cite{liu2026realwonder} combine traditional physics solvers with coarse video generation followed by appearance refinement, aiming to balance physical controllability with visual realism. NeuMA~\cite{cao2024neuma} and NewtonGen~\cite{yuan2025newtongen} instead explore neural physics simulation, replacing hand-crafted solvers with learned dynamics models that can be more naturally integrated into generative pipelines. Phys4DGen~\cite{lin2025phys4dgen}, PSIVG~\cite{foo2026physical}, and other methods~\cite{zhucp4d, li2026learning} further extend explicit physical modeling toward 4D simulation and visualization, enriching the representation of dynamic scenes beyond static geometry. Despite these advances, existing explicit simulation based approaches still depend on scene reconstruction, simulator coupling, or specialized physical states, which limits their scalability in open-domain video generation.

\subsection{Physics-informed video generation with implicit constraints}
Physics-informed video generation can also be pursued without explicit simulation by introducing implicit physical constraints into the generation process. Physical reasoning has been incorporated through chain-of-thought decomposition, reasoning-guided prompt refinement, and staged motion planning to steer synthesis toward physically plausible outcomes~\cite{xue2025phyt2v,zhao2026phyrpr,wang2026chain}. Physical supervision has also been imposed through representation alignment, preference optimization, reward-guided sampling, and world-model-based objectives to encourage physically consistent dynamics during training or inference~\cite{yuan2026inference,shen2026phantom, zhang2025videorepa,cai2025phygdpo}. Structured physical priors have further been introduced through mixture-of-experts architectures, which specialize different components for distinct physical principles or phenomena~\cite{wang2025wisa,wang2025prophy}.
However, these methods still impose physics indirectly through prompts, auxiliary objectives, or coarse expert priors, leaving limited control over how physical constraints are represented and propagated inside the generator. In contrast, our method constructs a shared operational physical interface together with category-specific operators, enabling structured constraint alignment directly in latent dynamics.

%% file: sec/3_method.tex
\section{Methodology}
\label{sec:method}
\paragraph{Overview.}
PILA follows a three-stage information flow. First, a shared physical encoder maps the latent state of a frozen flow-matching generator into a physical feature, which AFE converts into an operational physical attribute bank organized by fixed field-proxy slots, while an LLM-assisted router selects the relevant physical-category experts (Sec.~\ref{sec:physics_experts}). Second, the selected operator experts refine this bank through recipe-masked residual updates and category-specific operational residuals (Sec.~\ref{sec:physics_module}). Third, the refined physical attribute bank is decoded into a lightweight latent vector-field correction that aligns the pretrained latent vector field with the routed physical interface (Sec.~\ref{sec:physics_aligned_fm}).

\begin{figure*}[t]
    \centering
    \includegraphics[width=1\linewidth]{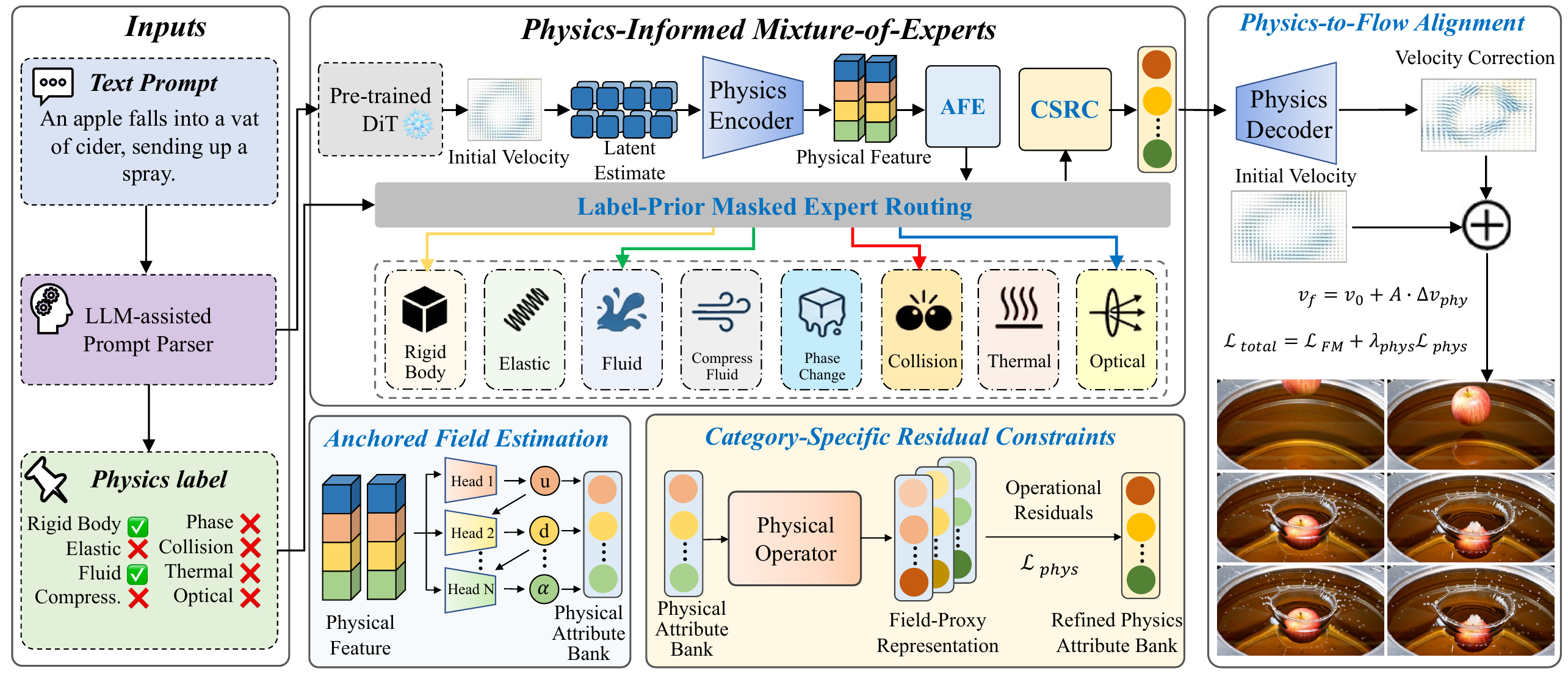}
    \caption{\textbf{Pipeline of PILA}. The method proceeds through three stages: (1) operational physical interface encoding and label-prior expert routing (Sec.~\ref{sec:physics_experts}); (2) physics-structured operator refinement of category-specific field-proxy representations (Sec.~\ref{sec:physics_module}); and (3) physics-to-latent flow alignment through a lightweight correction to the frozen flow-matching vector field (Sec.~\ref{sec:physics_aligned_fm}).}
    \label{fig:pipeline}
    \vspace{-0.4cm}
\end{figure*}

\subsection{Physical Interface Encoding and Expert Routing}
\label{sec:physics_experts}

Under the flow-matching formulation of the frozen video generator, $v_{\theta_0}(z_t,t)$ denotes the predicted latent velocity for the noisy latent $z_t$ at noise level $\sigma$.
We then derive a latent-space estimate $\hat{\mathbf{x}}_0$ as the input to the physical encoder:
\begin{equation}
    \hat{\mathbf{x}}_0
    =
    z_t-\sigma v_{\theta_0}(z_t,t).
    \label{eq:xphys_estimate}
\end{equation}
The fixed generation condition of the frozen backbone is omitted from $v_{\theta_0}$ for notational clarity.
This estimate is first encoded into a physical feature:
\begin{equation}
    \mathbf{f}_{\mathrm{phys}}
    =
    E_{\mathrm{phys}}(\hat{\mathbf{x}}_{0}, \sigma).
    \label{eq:physical_feature}
\end{equation}
The \textbf{anchored field estimation (AFE)} module then constructs the operational physical attribute bank.
\textcolor{black}{Let $\mathcal{Q}$ denote the set of field-proxy quantities represented by the bank, we obtain the operational physical attribute bank $\mathbf{a}$ by:}
\begin{equation}
    \mathbf{a}
    =
    \mathrm{AFE}(\mathbf{f}_{\mathrm{phys}})
    =
    [\mathbf{a}^{(q)}]_{q\in\mathcal{Q}}
    \in \mathbb{R}^{32\times T\times H\times W},
    \label{eq:attribute_bank}
\end{equation}
where $\mathbf{a}^{(q)}$ denotes the slot of $\mathbf{a}$ associated with quantity $q$.

The channels of $\mathbf{a}$ are organized into fixed field-proxy slots for displacement, velocity, pressure, density, temperature, phase/support, strain, stress, impulse, and wave-like activity.
Since these cues vary in their observability from video, AFE follows an observable-first construction: a motion-derived velocity proxy $\hat{u}$ serves as a kinematic anchor where applicable, while weakly observed variables are introduced progressively through quantity-specific heads.
We use ``kinematic anchor'' for this AFE-internal proxy to distinguish it from the flow-matching latent velocity $v_{\theta_0}$.
The resulting slots are operational field proxies used by routed constraints, not exact physical-state estimates.
Appendix Sec.~\ref{sec:afe_field_construction} details the field construction used by AFE.

We further introduce a \textbf{label-prior masked expert routing (LPMER)} strategy to select category-specific physics operators for the current sample.
An LLM-assisted prompt parser first normalizes the input prompt into a concise physics-oriented description and extracts a compact set of relevant physical-category labels.
The refined prompt description is encoded into a condition vector $\mathbf{c}$, while the extracted category labels are converted into label-prior logits $\mathbf{b}_{\mathrm{label}}$.
The router combines learned condition-dependent logits with these label priors to obtain expert routing scores.
To balance generation quality and computational efficiency, the model ranks experts by their routing scores and activates a predefined top-$k$ subset:
\begin{equation}
    \boldsymbol{\ell}=R_{\psi}(\mathbf{c})+\mathbf{b}_{\mathrm{label}},
    \qquad
    \mathcal{K}=\mathrm{TopK}(\boldsymbol{\ell}),
    \qquad
    w_k=\mathrm{softmax}(\boldsymbol{\ell}_{\mathcal{K}}/\tau)_k .
    \label{eq:operator_routing}
\end{equation}
The selected set $\mathcal{K}$ determines which operator experts may refine the shared physical interface, while the routing weights $\{w_k\}$ determine how their updates are combined downstream.
Appendix Sec.~\ref{sec:lpmer_routing_details} provides additional details on the LPMER routing construction.

\begin{figure*}[t]
    \centering
    \includegraphics[width=1\linewidth]{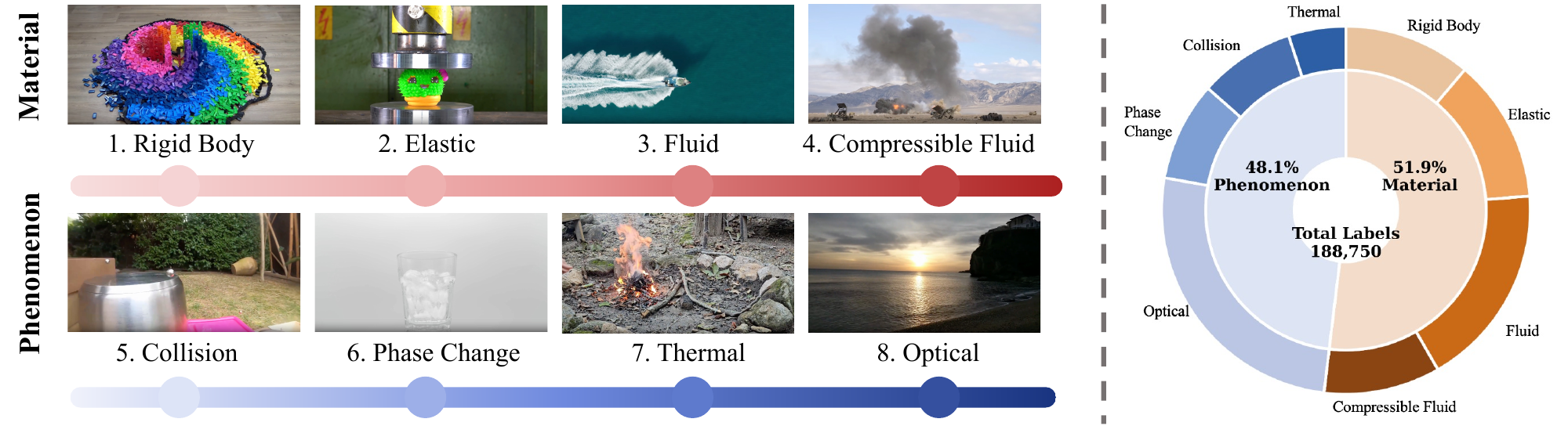}
    \caption{\textbf{Training data distribution across physical categories.} We relabel WISA-80K into eight physical categories, grouped into four material categories and four phenomenon categories.}
    \label{fig:data}
    \vspace{-0.4cm}
\end{figure*}

\begin{table}[t]
    \centering
    \caption{\textbf{Category-specific operational residual constraints.} For each expert, we list the field proxies and representative residual terms $\mathcal{C}_{k,i}$ used during training.}
    \footnotesize
    \setlength{\tabcolsep}{1pt}
    \renewcommand{\arraystretch}{1.0}
    \begin{tabularx}{\linewidth}{@{} l |l| X @{}}
    \toprule
    \textbf{Physical Category} & \textbf{Proxy Fields} & \textbf{Representative Operational Residuals } $\mathcal{C}_{k,i}$ \\
    \midrule
    1. Rigid Body & $d,u,\varepsilon,\sigma$ & $\partial_t d_k-u_k,\; \varepsilon_k,\; \sigma_k-\gamma_{\sigma}\varepsilon_k,\; \mathrm{sym}\nabla u_k$ \\

    2. Elastic & $d,u,\varepsilon,\sigma$ & $\partial_t^2 d_k-c_d^2\Delta d_k,\; \partial_t^2 u_k-c_u^2\Delta u_k,\; \varepsilon_k-\mathrm{sym}\nabla d_k,\; \sigma_k-\gamma_{\sigma}\varepsilon_k$ \\

    3. Fluid & $u,p,\rho$ & $\partial_t \rho_k+\nabla\!\cdot(\rho_k u_k),\; \partial_t u_k+(u_k\!\cdot\!\nabla)u_k+\nabla p_k/\rho_k-\nu\Delta u_k,\; \nabla p_k,\; \nabla \rho_k+\mathrm{ReLU}(\rho_0-\rho_k)$ \\

    4. Compress. & $u,p,\rho$ & $\partial_t \rho_k+\nabla\!\cdot(\rho_k u_k),\; \partial_t(\nabla\!\cdot u_k)-c_p\Delta p_k,\; p_k-\gamma_p\rho_k,\; \nabla u_k$ \\

    5. Phase Change & $u,\rho,T,\alpha$ & $\partial_t\alpha_k+\nabla\!\cdot(\alpha_k u_k),\; \partial_t T_k+u_k\!\cdot\nabla T_k-\kappa\Delta T_k-\beta\partial_t\alpha_k,\; \rho_k-\gamma_{\rho\alpha}\alpha_k,\; \nabla\alpha_k$ \\

    6. Collision & $d,u,j$ & $\partial_t d_k-u_k,\; \partial_t u_k-\gamma_j j_k,\; \nabla\!\cdot d_k+\bar{j}_k,\; \Delta j_k$ \\

    7. Thermal & $u,T$ & $\partial_t T_k+\nabla\!\cdot(u_k\,\bar{T}_k)-\kappa\Delta T_k,\; \nabla T_k$ \\

    8. Optical & $\psi,\alpha$ & $\partial_t^2 \psi_k-c_{\psi}^2\Delta \psi_k,\; (1-\alpha_k)\psi_k,\; \Delta \alpha_k$ \\
    \bottomrule
    \end{tabularx}
    \label{tab:category_constraints}
    \vspace{-0.4cm}
\end{table}

\begin{table}[t]
    \centering
    \caption{\textbf{Quantitative comparison on VideoPhy~\cite{bansal2024videophy} dataset.} Joint and Rule are evaluated using VideoPhy-2~\cite{bansal2025videophy}. Dynamic, Mechanics, Material, Motion are evaluated using VBench-2.0~\cite{zheng2025vbench}. Imaging Quality (Imaging) is evaluated using VBench~\cite{huang2024vbench}. The $^\dagger$ row transfers the Wan 2.1-trained adapter to Wan 2.2-14B without 14B-specific training. \textbf{Bold}: Best. \underline{Underline}: Second Best.}
    \setlength{\tabcolsep}{3pt}
    \renewcommand{\arraystretch}{1.0}
    \begin{tabular*}{\linewidth}{@{\extracolsep{\fill}} l | c c | c c c c | c @{}}
    \toprule
    \textbf{Method} & \textbf{Joint}↑ & \textbf{Rule}↑ & \textbf{Dynamic}↑ & \textbf{Mechanics}↑ & \textbf{Material}↑ & \textbf{Motion}↑ & \textbf{Imaging}↑ \\
    \midrule
    Wan 2.1-1.3B~\cite{wan2025wan} & 0.643 & 0.852 & 0.729 & 0.518 & 0.755 & 0.436 & \underline{0.589} \\
    CogVideoX-5B~\cite{yangcogvideox}  & \underline{0.716} & \underline{0.902} & \underline{0.776} & 0.520 & \textbf{0.887} & 0.459 & 0.554 \\
    VideoREPA-2B~\cite{zhang2025videorepa} & 0.684 & 0.882 & 0.773 & \underline{0.521} & 0.790 & \underline{0.474} & 0.519 \\
    \textbf{Ours-1.3B} & \textbf{0.740} & \textbf{0.965} & \textbf{0.933} & \textbf{0.542} & \underline{0.873} & \textbf{0.576} & \textbf{0.600} \\
    \midrule
    Hunyuan-80G~\cite{kong2024hunyuanvideo} & 0.546 & 0.829 & 0.616 & 0.551 & 0.793 & 0.337 & 0.637 \\
    Wan 2.2-14B~\cite{wan2025wan} & \underline{0.690} & \underline{0.926} & \underline{0.823} & \underline{0.563} & \underline{0.842} & \underline{0.552} & \underline{0.653} \\
    WISA-14B~\cite{wang2025wisa} & 0.681 & 0.897 & 0.794 & 0.519 & 0.820 & 0.517 & 0.626 \\
    \textbf{Ours-14B$^\dagger$} & \textbf{0.743} & \textbf{0.953} & \textbf{0.953} & \textbf{0.608} & \textbf{0.907} & \textbf{0.651} & \textbf{0.658} \\
    \bottomrule
    \end{tabular*}
    \vspace{-0.5cm}
    \label{tab:quant}
\end{table}

\subsection{Physics-Structured Residual Modeling}
\label{sec:physics_module}

Different physical categories in our taxonomy are associated with different forms of residual structure, such as PDE-style anchors, kinematic consistency terms, latent closure proxies, and stabilizing priors.
We therefore introduce \textbf{category-specific residual constraints (CSRC)}, which regularize the operational physical attribute bank through residual families associated with the routed categories.
Scale factors such as $c_d$, $c_{\psi}$, and $\gamma_{\sigma}$ are latent-space normalizers, not measured material parameters.
Each selected expert is implemented as an operator over the shared physical interface: it receives the condition vector $\mathbf{c}$ and the current bank $\mathbf{a}$, then predicts a residual refinement in the same attribute space.
A recipe mask $\mathbf{M}_k$ restricts the refinement to the field slots associated with category $k$:
\begin{equation}
    \Delta \mathbf{a}_k
    =
    \mathbf{M}_k \odot O_k([\mathbf{c},\mathbf{a}]),
    \qquad
    \boldsymbol{\phi}_k
    =
    P_k(\mathbf{a}+\Delta\mathbf{a}_{k}).
    \label{eq:masked_attribute_update}
\end{equation}
Here, $O_k$ is the learnable operator expert for category $k$, and $P_k$ is a category-specific field readout that extracts the proxy variables required by the corresponding residual constraints. The resulting $\boldsymbol{\phi}_k$ denotes a category-specific field-proxy representation, obtained by reading out the category-relevant slots from the expert-refined physical attribute bank.

The residual objective for expert $k$ is then formulated as:
\begin{equation}
    \mathcal{R}_{k}(\boldsymbol{\phi}_k)
    =
    \sum_{i} s_{k,i}\,
    \big\|
    \mathcal{C}_{k,i}
    (\boldsymbol{\phi}_k)
    \big\|^2 ,
    \label{eq:physics_residual1}
\end{equation}
where $\mathcal{C}_{k,i}$ denotes the $i$-th differentiable residual term for category $k$, and $s_{k,i}$ is its adaptive weight.
Table~\ref{tab:category_constraints} summarizes the field proxies and representative residual families associated with each operator expert.
Appendix Sec.~\ref{sec:constraint_rationale} further categorizes these terms as PDE-style anchors, kinematic consistency terms, latent closure proxies, or stabilizing priors, and states the reference relation or prior behind each residual family.

For a sample $b$, let $\mathcal{K}_b$ be the top-$k$ expert set selected by the router, $\mathcal{Y}_b$ be the set of active physical-category labels predicted from the input prompt, and $\boldsymbol{\phi}_{b,k}$ be the corresponding category-specific field-proxy representation.
The routed residual objective is computed only over experts that are both selected and label-consistent:
\begin{equation}
    \mathcal{L}_{\mathrm{phys}}
    =
    \mathbb{E}_{b}\Bigg[
    \frac{1}{Z_b}
    \sum_{k\in \mathcal{K}_b\cap\mathcal{Y}_b}
    w_{b,k}\,
    \mathcal{R}_k(\boldsymbol{\phi}_{b,k})
    \Bigg],
    \label{eq:physics_residual}
\end{equation}
where
$Z_b=\sum_{k\in \mathcal{K}_b\cap\mathcal{Y}_b}w_{b,k}$
normalizes the active routing weights.
After applying operational residual constraints to the category-specific field-proxy representations, their residual refinements are aggregated into a refined physical attribute bank:
\begin{equation}
    \tilde{\mathbf{a}}_b
    =
    \mathbf{a}_b+\sum_{k\in\mathcal{K}_b}w_{b,k}\,\Delta\mathbf{a}_{b,k}.
    \label{eq:refined_attribute_bank}
\end{equation}
This refined physical attribute bank preserves the shared physical interface while incorporating the category-specific corrections selected by LPMER.
Appendix Sec.~\ref{sec:constraint_rationale} provides a term-level interpretation of the residual constraints.

\subsection{Physics-to-Latent Flow Alignment}
\label{sec:physics_aligned_fm}

Let $v_{\theta_0}(z_t,t)$ denote a pretrained flow-matching video generator, parameterized as a latent-space vector field.
Our goal is to align the frozen generative dynamics with the routed physical interface while preserving the visual prior of the pretrained backbone.

\paragraph{Attribute-to-flow correction.}
At each denoising step, the frozen backbone first predicts a base vector field $v_{\theta_0}(z_t,t)$.
The physics pathway constructs the latent estimate $\hat{\mathbf{x}}_0$, encodes it into the physical feature $\mathbf{f}_{\mathrm{phys}}$, converts the feature into the operational physical attribute bank $\mathbf{a}$ through AFE, and applies routed operator refinements to obtain the refined physical attribute bank $\tilde{\mathbf{a}}$.
To transfer this physics-aware representation back to the generative dynamics, we introduce a lightweight attribute-to-flow decoder $D_{\psi}$.

The decoder takes the refined physical attribute bank $\tilde{\mathbf{a}}$ as the primary physical input, the task condition vector $\mathbf{c}$ as semantic guidance, and the frozen backbone prediction $v_{\theta_0}(z_t,t)$ as the reference flow direction.
The condition vector is broadcast to the spatiotemporal latent resolution and concatenated with the refined physical attribute bank and the base vector field.
A small spatiotemporal projection network then maps this refined physical attribute bank to a residual vector field with the same shape as the backbone prediction:
\begin{equation}
    \Delta v_{\psi}
    =
    D_{\psi}\!\left([\mathbf{c},\tilde{\mathbf{a}},v_{\theta_0}(z_t,t)]\right).
    \label{eq:attribute_decoder}
\end{equation}
Here, $D_{\psi}$ does not decode physical attributes into pixels or explicit physical fields; instead, it converts the refined attribute representation into a latent-space correction that is compatible with the flow-matching vector field.

The physics-aligned vector field is then defined by residual composition:
\begin{equation}
    v_{\theta_0,\psi}(z_t,t)
    =
    v_{\theta_0}(z_t,t)
    +
    \Delta v_{\psi}.
    \label{eq:phys_conditioned_fm}
\end{equation}
The pretrained parameters $\theta_0$ are kept frozen, and only the physics adapter parameters $\psi$ are optimized.
Thus, physics-structured residual constraints affect generation through an additive latent correction, rather than by modifying the pretrained appearance model itself.

\paragraph{Training objective.}
The full training objective combines flow-matching supervision and routed operational residual constraints:
\begin{equation}
    \mathcal{L}_{\mathrm{total}}
    =
    \mathcal{L}_{\mathrm{FM}}
    +
    \lambda_{\mathrm{phys}}\mathcal{L}_{\mathrm{phys}},
    \label{eq:total_loss}
\end{equation}
where $\mathcal{L}_{\mathrm{phys}}$ is the routed operational residual loss defined in Sec.~\ref{sec:physics_module}.

%% file: sec/4_exp.tex
\section{Experiments}
\label{sec:exp}

\subsection{Experimental Setup}

\begin{figure*}[t]
    \centering
    \includegraphics[width=1\linewidth]{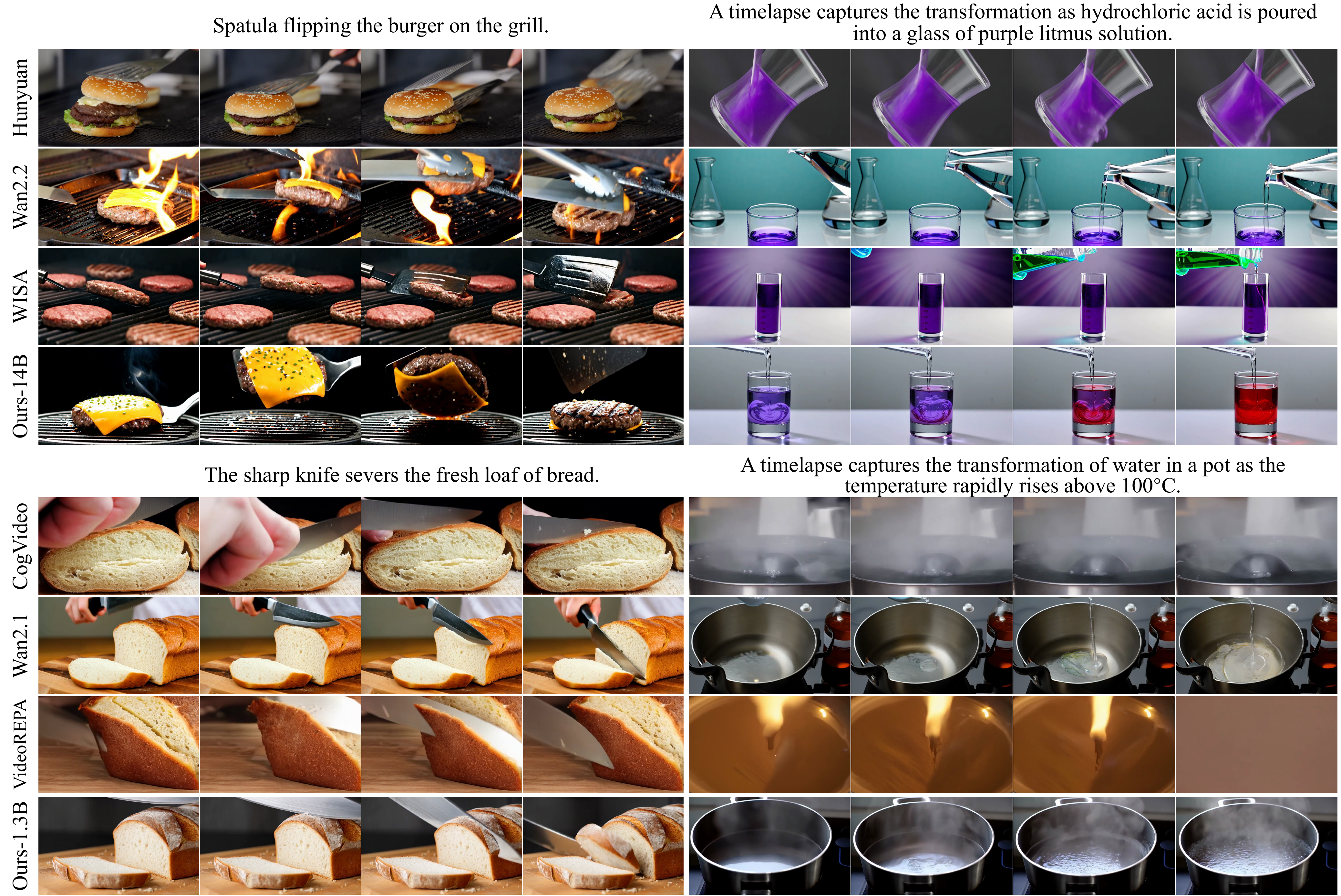}
    \caption{\textbf{Qualitative comparison with baselines.} PILA better captures physically grounded interactions and state changes across diverse phenomena.}
    \label{fig:com}
    \vspace{-0.5cm}
\end{figure*}

\begin{table}[t]
    \centering
    \caption{\textbf{Quantitative comparison with baselines evaluated on PhyGenBench~\cite{meng2024towards} dataset.} Mechanics, Optics, Thermal, Material and Average are evaluated using PhyGenBench~\cite{meng2024towards}. Motion is evaluated using VBench-2.0~\cite{zheng2025vbench}. Imaging Quality (Imaging) is evaluated using VBench~\cite{huang2024vbench}.}
    \setlength{\tabcolsep}{1.5pt}
    \renewcommand{\arraystretch}{1.0}
    \begin{tabular*}{\linewidth}{@{\extracolsep{\fill}} l | c c c c c | c | c @{}}
    \toprule
    \textbf{Method} & \textbf{Mechanics}↑ & \textbf{Optics}↑ & \textbf{Thermal}↑ & \textbf{Material}↑ & \textbf{Average}↑ & \textbf{Motion}↑ & \textbf{Imaging}↑ \\
    \midrule
    Wan 2.1-1.3B~\cite{wan2025wan} & \underline{0.483} & 0.640 & 0.322 & 0.342 & 0.467 & 0.250 & \underline{0.538}  \\
    CogVideoX-5B~\cite{yangcogvideox} & 0.483 & \underline{0.673} & 0.433 & \underline{0.425} & \underline{0.519} & 0.263 & 0.508 \\
    VideoREPA-2B~\cite{zhang2025videorepa} & 0.433 & 0.627 & \underline{0.433} & 0.417 & 0.490 & \underline{0.269} & 0.447  \\
    \textbf{Ours-1.3B} & \textbf{0.575} & \textbf{0.713} & \textbf{0.578} & \textbf{0.558} & \textbf{0.615} & \textbf{0.356} & \textbf{0.543} \\
    \midrule
    Hunyuan-80G~\cite{kong2024hunyuanvideo} & 0.375 & 0.420 & 0.211 & 0.325 & 0.346 & 0.150 & \textbf{0.574}  \\
    Wan 2.2-14B~\cite{wan2025wan} & 0.458 & \underline{0.753} & \underline{0.444} & \underline{0.417} & \underline{0.538} & 0.288 & 0.559 \\
    WISA-14B~\cite{wang2025wisa} & \underline{0.458} & 0.700 & 0.356 & 0.375 & 0.494 & \underline{0.306} & 0.535 \\
    \textbf{Ours-14B$^\dagger$} & \textbf{0.625} & \textbf{0.840} & \textbf{0.589} & \textbf{0.617} & \textbf{0.683} & \textbf{0.381} & \underline{0.568} \\
    \bottomrule
    \end{tabular*}
    \label{tab:quant2}
    \vspace{-0.4cm}
\end{table}

\noindent\textbf{Evaluation Dataset.}
We evaluate on two physics-focused benchmarks, VideoPhy~\cite{bansal2024videophy} and PhyGenBench~\cite{meng2024towards}. We sample 344 prompts from VideoPhy~\cite{bansal2024videophy}, covering solid-solid, solid-fluid, and fluid-fluid interactions, to evaluate whether generated videos remain physically plausible. PhyGenBench complements this setting with 160 prompts covering 27 physical laws across mechanics, optics, thermal, and material domains, together with domain-wise physics evaluation.

\noindent\textbf{Metrics.}
To evaluate physical plausibility, we adopt metrics from VideoPhy-2~\cite{bansal2025videophy}, VBench-2.0~\cite{zheng2025vbench}, and PhyGenBench~\cite{meng2024towards}.
VideoPhy-2~\cite{bansal2025videophy} reports a Joint Score (Joint), defined as the proportion of samples whose Semantic Adherence and Physical Commonsense are both greater than or equal to 4. It also evaluates Physical Rules (Rule) by determining whether a video violates predefined physical constraints.
For VBench-2.0~\cite{zheng2025vbench}, we adopt Dynamic Spatial Relationship (Dynamic) to assess entity controllability, Mechanics and Material to measure physical plausibility, and Motion Rationality (Motion) to evaluate commonsense reasoning.
For PhyGenBench~\cite{meng2024towards}, we utilize all available metrics, including Mechanics, Optics, Thermal, and Material.
In addition to physics evaluation, we include the Imaging Quality (Imaging) metric from VBench~\cite{huang2024vbench} to assess visual quality.

\subsection{Implementation Details}

We train PILA on top of Wan 2.1-T2V-1.3B~\cite{wan2025wan}. The pretrained video generator is frozen, and only the PILA adapter is optimized on WISA-80K~\cite{wang2025wisa} videos with standardized physical-category labels aligned to our four material and four phenomenon categories. Unless otherwise specified, training uses 81-frame videos at $480\times832$ resolution, a 32-channel operational physical attribute bank, top-$4$ routed category experts, finite-difference latent residuals, and \texttt{bf16} multi-GPU optimization. For Wan 2.2-T2V-A14B evaluation, we do not train a separate adapter; instead, we load the Wan 2.1-trained PILA adapter and apply it as a frozen correction module to the frozen Wan 2.2 backbone. Full training and transfer details are provided in Appendix Sec.~\ref{sec:appendix_implementation_details}.

\subsection{Comparison with State-of-the-Art}

\begin{table}[t]
    \centering
    \caption{\textbf{Ablation study on Wan 2.1-1.3B.} All variants use the same backbone and training budget.}
    \setlength{\tabcolsep}{1pt}
    \renewcommand{\arraystretch}{1.0}
    \begin{tabular*}{\linewidth}{@{\extracolsep{\fill}} l | c c c c c | c | c @{}}
    \toprule
    \textbf{Configuration} & \textbf{Mechanics}↑ & \textbf{Optics}↑ & \textbf{Thermal}↑ & \textbf{Material}↑ & \textbf{Average}↑ & \textbf{Imaging}↑ & \textbf{Motion}↑\\
    \midrule
    Baseline & 0.483 & 0.640 & 0.322 & 0.342 & 0.467 & 0.538 & 0.250  \\
    + Plain Adapter & 0.471 & 0.653 & 0.502 & 0.521 & 0.546 & 0.534 & 0.315 \\
    \midrule
    w/o Flow correction   & 0.455 & 0.563 & 	0.541 & 0.463 & 0.507 & 0.539  & 0.300 \\
    w/o AFE & 0.536 & \underline{0.693} & 0.486 & 0.395 & 0.540 &  0.537 & 0.325 \\
    w/o MoE operator  & 0.550 & 0.637 & \underline{0.572} & 0.536 & 0.578 & 0.538  & 0.334 \\
    w/o Residual Constraints & 0.548 & 0.685 & 0.531 & 0.523 & 0.582 & \underline{0.542} & 0.344 \\
    w/o Prompt Refine & 0.562 & 0.668 &
    0.555 & \underline{0.549} & 0.591 & 0.540  & \underline{0.352} \\
    w/o Expert Routing & \underline{0.570} & 0.681 & 0.571 & 0.538 & \underline{0.597} & 0.541  & 0.347 \\
    \midrule
    Ours-1.3B (Full) & \textbf{0.575} & \textbf{0.713} & \textbf{0.578} & \textbf{0.558} & \textbf{0.615} & \textbf{0.543}  & \textbf{0.356}  \\
    \bottomrule
    \end{tabular*}
    \vspace{-0.5cm}
    \label{tab:ablation}
\end{table}

\begin{figure}[t]
    \centering
    \includegraphics[width=0.95\linewidth]{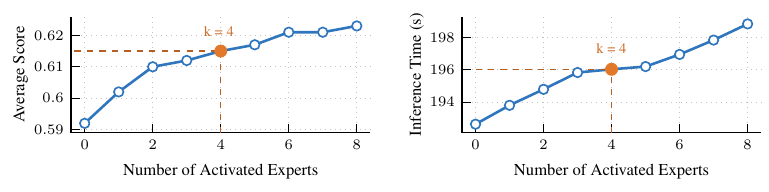}
    \vspace{-0.3cm}
    \caption{\textbf{Quality-efficiency trade-off with different numbers of routed experts.} Orange markers denote the default $k=4$ setting.}
    \label{fig:expert_tradeoff}
    \vspace{-0.5cm}
\end{figure}

We compare PILA with recent strong video-generation baselines, including Hunyuan Video~\cite{kong2024hunyuanvideo}, CogVideoX~\cite{yangcogvideox}, Wan~\cite{wan2025wan}, VideoREPA~\cite{zhang2025videorepa}, and WISA~\cite{wang2025wisa}. We report 1.3B--5B and 14B-scale results separately, with Ours-14B$^\dagger$ denoting direct transfer of the Wan 2.1-trained adapter to Wan 2.2-14B without additional adapter optimization.
Tables~\ref{tab:quant} and~\ref{tab:quant2} show consistent gains in physical plausibility. On VideoPhy-2, Ours-1.3B improves Joint and Rule by $15.1\%$ and $13.3\%$ over its Wan backbone, respectively. Under the 14B transfer setting, PILA improves all physics-oriented VideoPhy-2~\cite{bansal2025videophy} and VBench-2.0~\cite{zheng2025vbench} metrics over Wan 2.2-14B, with the largest gains on Motion and Dynamic. On PhyGenBench~\cite{meng2024towards}, PILA improves the average score by $18.5\%$ over the strongest 1.3B--5B baseline and by $27.0\%$ over the strongest 14B baseline. Qualitative comparisons in Figure~\ref{fig:com} further show better object--tool interaction, acid--base color transition, cutting-induced shape change, and boiling dynamics than baselines.

\subsection{Ablation Studies}

\paragraph{Main component ablations.} Table~\ref{tab:ablation} shows that PILA is not merely a capacity effect, as the full model improves the average physics score by $12.6\%$ over the capacity-matched plain adapter. The largest degradation comes from removing the physics-to-flow correction, indicating that the physical bank is most effective when it directly modulates the FM vector field. The drop caused by removing the AFE module further shows that stable physical modeling requires a dedicated latent-to-physics feature interface. 
More detailed explanation and additional ablations are listed in Appendix Sec.~\ref{sec:appendix_ablation_details}. 

\paragraph{Quality-Efficiency Trade-off Study.}

Figure~\ref{fig:expert_tradeoff} shows that activating more experts improves the average physics-oriented score, but the gain becomes small after the default setting. Moving from $k=4$ to $k=8$ adds only 0.008 score while increasing latency by 2.81s, so we use $k=4$ as a balanced configuration in all main experiments.

\begin{figure*}
    \centering
    \includegraphics[width=1\linewidth]{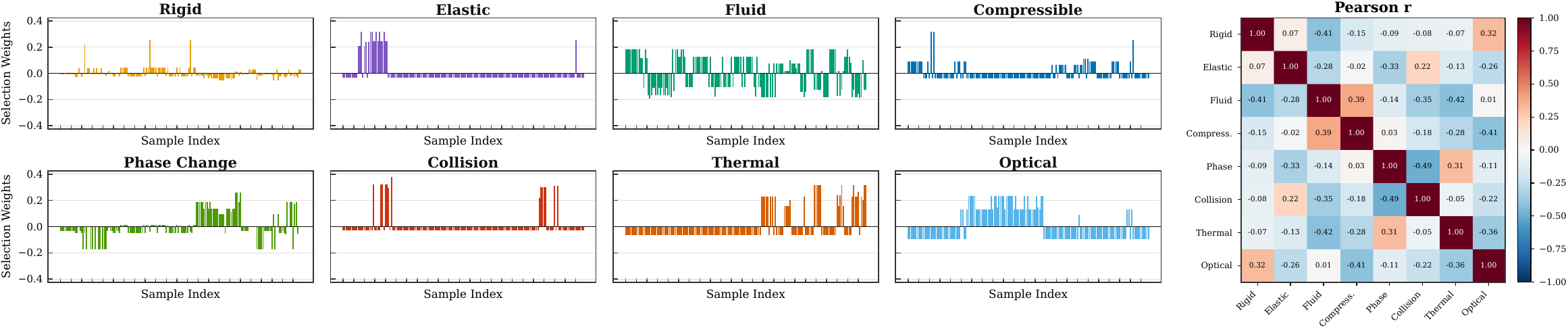}
    \vspace{-0.6cm}
    \caption{\textbf{Analysis of the expert router}. $r$ represents the Pearson correlation coefficient calculated between different distributions.}
    \label{fig:router}
    \vspace{-0.4cm}
\end{figure*}

\begin{figure}
    \centering
    \includegraphics[width=1\linewidth]{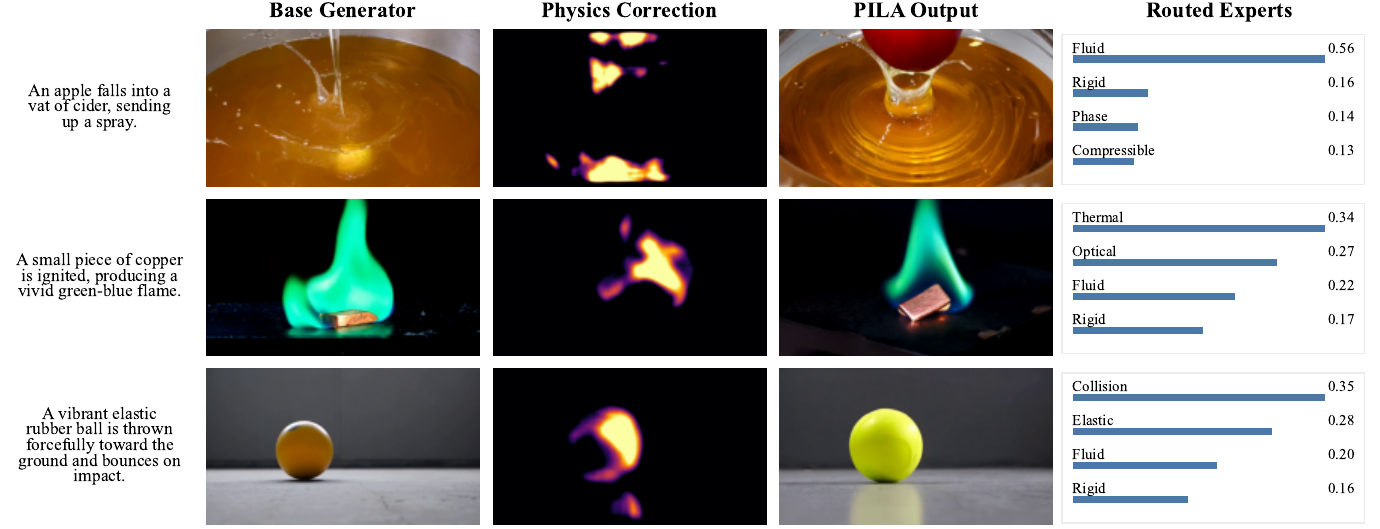}
    \caption{\textbf{Visualization of routed expert activations.} Expert maps show that PILA applies category-specific updates to physically active regions.}
    \label{fig:expert_map}
    \vspace{-0.5cm}
\end{figure}

\subsection{Analysis of Learned Components}

To better understand how PILA internally organizes physical information, we analyze the behavior of its routed experts. Figure~\ref{fig:expert_map} visualizes the spatial activation of different experts, showing that the routed branch concentrates corrections on physically active regions rather than applying a uniform global modification. Figure~\ref{fig:router} further summarizes the router statistics, where different experts exhibit category-specific activation patterns, and physically related categories, such as Thermal and Phase Change, show stronger correlations, consistent with their natural coupling in real-world dynamics.

%% file: sec/5_conclu.tex
\section{Conclusion}
\label{sec:con}
In this work, we presented PILA, a multi-expert physical latent-space constraint alignment framework for integrating physics-aware guidance into pretrained flow-matching video generators. PILA constructs an operational physical attribute bank from generator latents through anchored field estimation, and uses label-prior masked expert routing to activate category-relevant physical experts with prompt-derived priors while retaining adaptive routing. Each routed expert reads the corresponding field-proxy representation from the attribute bank and applies category-specific operational residual constraints, after which the refined bank provides a lightweight correction signal for the flow-matching vector field. Trained on Wan 2.1-1.3B and directly transferred to Wan 2.2-14B, PILA improves motion continuity and benchmark-measured physical plausibility while preserving visual quality. Experiments on VBench-2.0, VideoPhy-2, and PhyGenBench demonstrate state-of-the-art performance, suggesting that operational physical attribute modeling and routed physics-structured alignment provide an effective path toward more physically plausible video generation without requiring calibrated simulator states.

%% file: sec/X_suppl.tex
\clearpage
\setcounter{page}{1}
\appendix
\section{Additional Method Details}
\label{sec:appendix}

\subsection{AFE Field Construction}
\label{sec:afe_field_construction}

This section details how AFE converts the encoder feature into the operational physical attribute bank used by the routed constraints. The construction is intended as an operational latent interface: its slots are field proxies shaped by observable motion and operational residuals, not recovered physical states.

The attribute bank follows a fixed 32-channel contract. Let
\begin{equation}
    \mathcal{Q}
    =
    (d,u,p,\rho,T,\alpha,\varepsilon,\sigma,j,\psi),
    \qquad
    \mathbf{c}_{\mathcal{Q}}
    =
    (4,4,2,2,2,2,4,4,4,4),
    \label{eq:appendix_bank_contract}
\end{equation}
where $\mathbf{c}_{\mathcal{Q}}$ gives the channel width of each slot. The slots correspond to displacement ($d$), velocity ($u$), pressure ($p$), density ($\rho$), temperature ($T$), phase ($\alpha$), strain ($\varepsilon$), stress ($\sigma$), impulse ($j$), and wave ($\psi$). The encoder first produces $\mathbf{f}_{\mathrm{phys}}$, and a scaffold head maps it to a pre-AFE slot scaffold:
\begin{equation}
    \bar{\mathbf{a}}
    =
    H_{\mathrm{scaf}}(\mathbf{f}_{\mathrm{phys}}),
    \qquad
    \mathbf{a}
    =
    \mathrm{AFE}(\bar{\mathbf{a}},\mathbf{f}_{\mathrm{phys}})
    =
    [\mathbf{a}^{(q)}]_{q\in\mathcal{Q}} .
    \label{eq:appendix_afe_bank}
\end{equation}
Here, $\bar{\mathbf{a}}$ is an internal implementation scaffold, while the post-AFE bank $\mathbf{a}$ serves as the shared physical interface used throughout the method. The final bank is obtained by reading out a motion-derived velocity proxy as a kinematic anchor, constructing derived kinematic fields, and progressively activating weakly observed slots before repacking the same contract.

\noindent\textbf{Kinematic anchor and descriptors.}
AFE uses a motion-derived velocity proxy as the primary kinematic anchor where applicable. This terminology separates the AFE-internal proxy $\hat{u}$ from the flow-matching latent velocity $v_{\theta_0}$. The velocity scaffold slot is 4-channel, and a lightweight head reads it into a 2-channel velocity proxy
\begin{equation}
    \hat{u}
    =
    H_u(\bar{\mathbf{a}}^{(u)})
    =
    (\hat{u}_x,\hat{u}_y).
    \label{eq:appendix_velocity_readout}
\end{equation}
Finite differences on $\hat{u}$ then define the kinematic descriptors used by downstream field constructors:
\begin{equation}
    \nabla\!\cdot\hat{u}
    =
    \partial_x\hat{u}_x+\partial_y\hat{u}_y,
    \qquad
    \nabla\!\times\hat{u}
    =
    \partial_x\hat{u}_y-\partial_y\hat{u}_x,
    \label{eq:appendix_velocity_divcurl}
\end{equation}
\begin{equation}
    \hat{\varepsilon}
    =
    \left(
    \partial_x\hat{u}_x,\;
    \frac{1}{2}(\partial_y\hat{u}_x+\partial_x\hat{u}_y),\;
    \frac{1}{2}(\partial_y\hat{u}_x+\partial_x\hat{u}_y),\;
    \partial_y\hat{u}_y
    \right).
    \label{eq:appendix_strain_descriptor}
\end{equation}
These descriptors are finite-difference operational fields; they provide local motion geometry rather than exact strain measurements.

\noindent\textbf{Pressure, density, displacement, and stress.}
Pressure ($p$) and density ($\rho$) are constructed from the physical feature and kinematic-anchor descriptors:
\begin{equation}
    (\tilde{p},\tilde{\rho})
    =
    C_{p\rho}\!\left(
    [\mathbf{f}_{\mathrm{phys}},\hat{u},\|\hat{u}\|,
    \nabla\!\cdot\hat{u},\nabla\!\times\hat{u}]
    \right),
    \label{eq:appendix_prho_constructor}
\end{equation}
\begin{equation}
    \hat{p}
    =
    \tilde{p}-\mathrm{mean}(\tilde{p}),
    \qquad
    \hat{\rho}
    =
    \mathrm{softplus}(\tilde{\rho})+\rho_{\min}.
    \label{eq:appendix_prho_normalization}
\end{equation}
The pressure proxy is centered per sample to remove arbitrary offsets, while the density proxy is mapped to a positive range using a small floor $\rho_{\min}$.

Displacement ($d$) is constructed by temporal integration of the velocity proxy around the middle frame $t_0=\lfloor T/2\rfloor$. With $\hat{d}_{t_0}=0$, AFE applies midpoint integration forward and backward:
\begin{equation}
    \hat{d}_{t}
    =
    \hat{d}_{t-1}
    +
    \Delta t_{t-1}\,
    \frac{\hat{u}_{t-1}+\hat{u}_{t}}{2},
    \quad t>t_0,
    \qquad
    \hat{d}_{t}
    =
    \hat{d}_{t+1}
    -
    \Delta t_{t}\,
    \frac{\hat{u}_{t}+\hat{u}_{t+1}}{2},
    \quad t<t_0 .
    \label{eq:appendix_displacement_integration}
\end{equation}
The resulting 2-channel displacement proxy is repeated to fill the 4-channel displacement slot.

The stress slot is a closure proxy built from centered pressure and the strain-like descriptor. Let $\mathrm{tr}(\hat{\varepsilon})=\hat{\varepsilon}_{xx}+\hat{\varepsilon}_{yy}$. AFE uses
\begin{equation}
\begin{aligned}
    \hat{\sigma}_{xx}
    &=
    -\hat{p}+2\mu\hat{\varepsilon}_{xx}
    +\lambda\,\mathrm{tr}(\hat{\varepsilon}),\\
    \hat{\sigma}_{xy}
    &=
    2\mu\hat{\varepsilon}_{xy},
    \qquad
    \hat{\sigma}_{yx}
    =
    2\mu\hat{\varepsilon}_{yx},\\
    \hat{\sigma}_{yy}
    &=
    -\hat{p}+2\mu\hat{\varepsilon}_{yy}
    +\lambda\,\mathrm{tr}(\hat{\varepsilon}).
\end{aligned}
    \label{eq:appendix_stress_proxy}
\end{equation}
Here, $\mu$ and $\lambda$ are fixed scale factors in the latent proxy space. This relation is used only to provide a stress-like operational field compatible with downstream residuals.

\noindent\textbf{Progressive recovery of weakly observed slots.}
Weakly observed variables are introduced progressively. In the only-u recovery policy, the active set grows as
\[
    \texttt{core}
    \rightarrow
    \alpha
    \rightarrow
    T
    \rightarrow
    j
    \rightarrow
    \psi .
\]
Table~\ref{tab:afe_field_constructors} summarizes the constructors. Scalars are repeated to the channel width required by the bank contract, and inactive fields are zeroed before the final contract is packed.

\begin{table}[t]
    \centering
    \scriptsize
    \setlength{\tabcolsep}{2.5pt}
    \renewcommand{\arraystretch}{1.06}
    \caption{\textbf{AFE field constructors.} Each constructor produces an operational field proxy that is repacked into the fixed attribute-bank contract.}
    \label{tab:afe_field_constructors}
    \begin{tabularx}{\linewidth}{@{} l X X c @{}}
    \toprule
    \textbf{Slot (meaning)} & \textbf{Constructor} & \textbf{Context} & \textbf{Phase} \\
    \midrule
    $u$ (velocity) & $\hat{u}=H_u(\bar{\mathbf{a}}^{(u)})$ & 4-channel velocity scaffold slot & \texttt{core} \\
    $p,\rho$ (pressure, density) & $(\hat{p},\hat{\rho})$ from Eqs.~\ref{eq:appendix_prho_constructor}--\ref{eq:appendix_prho_normalization} & $\mathbf{f}_{\mathrm{phys}}$, $\hat{u}$, $\|\hat{u}\|$, divergence, curl & \texttt{core} \\
    $\varepsilon$ (strain) & $\hat{\varepsilon}$ from Eq.~\ref{eq:appendix_strain_descriptor} & spatial finite differences of $\hat{u}$ & \texttt{core} \\
    $\sigma$ (stress) & $\hat{\sigma}$ from Eq.~\ref{eq:appendix_stress_proxy} & $\hat{p}$ and $\hat{\varepsilon}$ & \texttt{core} \\
    $d$ (displacement) & $\hat{d}$ from Eq.~\ref{eq:appendix_displacement_integration} & midpoint temporal integration of $\hat{u}$ & \texttt{core} \\
    $\alpha$ (phase/support) & $\hat{\alpha}=\mathrm{sigmoid}(H_{\alpha}([\mathbf{f}_{\mathrm{phys}},\hat{u},\hat{\rho},\hat{d}]))$ & motion, density, displacement & $\alpha$ \\
    $T$ (temperature) & $\hat{T}=H_T([\mathbf{f}_{\mathrm{phys}},\hat{u},\hat{\rho},\hat{\alpha}])$ & motion, density, phase & $T$ \\
    $j$ (impulse) & $\hat{j}=H_j([\mathbf{f}_{\mathrm{phys}},\hat{d},\hat{u},\hat{p},\hat{\rho},\hat{\varepsilon},\hat{\sigma}])$ & kinematic and closure proxies & $j$ \\
    $\psi$ (wave) & $\hat{\psi}=H_{\psi}([\mathbf{f}_{\mathrm{phys}},\hat{\alpha}])$ & physical feature and phase/support proxy; repeated to the 4-channel wave slot & $\psi$ \\
    \bottomrule
    \end{tabularx}
\end{table}

After all active constructors have been evaluated, AFE repacks the constructed field proxies into the fixed slot order in Eq.~\ref{eq:appendix_bank_contract}. Fields that are not active in the current recovery phase are explicitly zeroed under the only-u policy. The constructed field proxies are later consumed by CSRC; Sec.~\ref{sec:constraint_rationale} explains how the routed residuals should be interpreted.

\subsection{LPMER Routing Details}
\label{sec:lpmer_routing_details}

LPMER connects prompt-level physical semantics to category-specific operators without making routing a fixed single-label decision. We use the physical-category set
\[
\begin{aligned}
    \mathcal{P}
    =
    \{&
    \text{Rigid Body},\text{Elastic},\text{Fluid},\text{Compressible Flow},\\
    &\text{Phase Change},\text{Collision},\text{Thermal},\text{Optical}
    \}
\end{aligned}
\]
where $\mathcal{P}$ denotes the eight active physical-category experts used in the paper. For each sample, the available category evidence is converted into a multi-label set $\mathcal{Y}\subseteq\mathcal{P}$. During training, $\mathcal{Y}$ is read from the relabeled WISA-80K annotations described in Sec.~\ref{sec:dataset_taxonomy}; during inference, the same metadata format can be produced by the LLM-assisted prompt parser after normalizing the prompt into a concise physics-oriented description. This description provides the condition vector $\mathbf{c}$, while the extracted labels define the routing prior. The parser does not supervise the constructed field proxies.

\begin{figure*}[ht]
    \centering
    \includegraphics[width=1\linewidth]{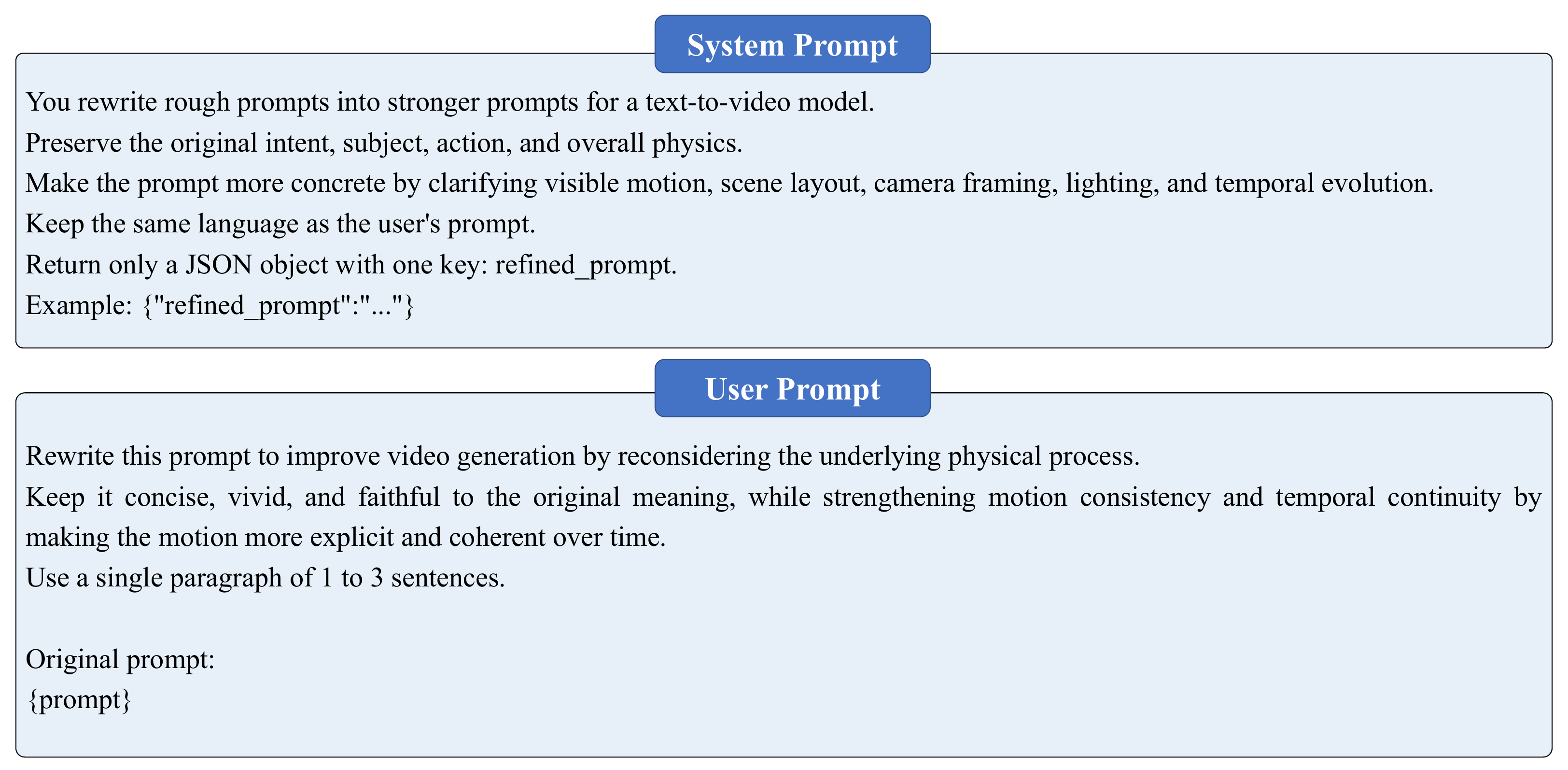}
    \caption{\textbf{Prompt-refinement template used before LPMER routing.} The template rewrites a raw input prompt into a concise physics-oriented description that emphasizes objects, motion, interactions, material cues, and state changes. This refined description is used as cleaner prompt evidence for the subsequent category-labeling and label-prior routing steps.}
    \vspace{-0.3cm}
    \label{fig:lpmer_prompt_refine}
\end{figure*}

\begin{figure*}[ht]
    \centering
    \includegraphics[width=1\linewidth]{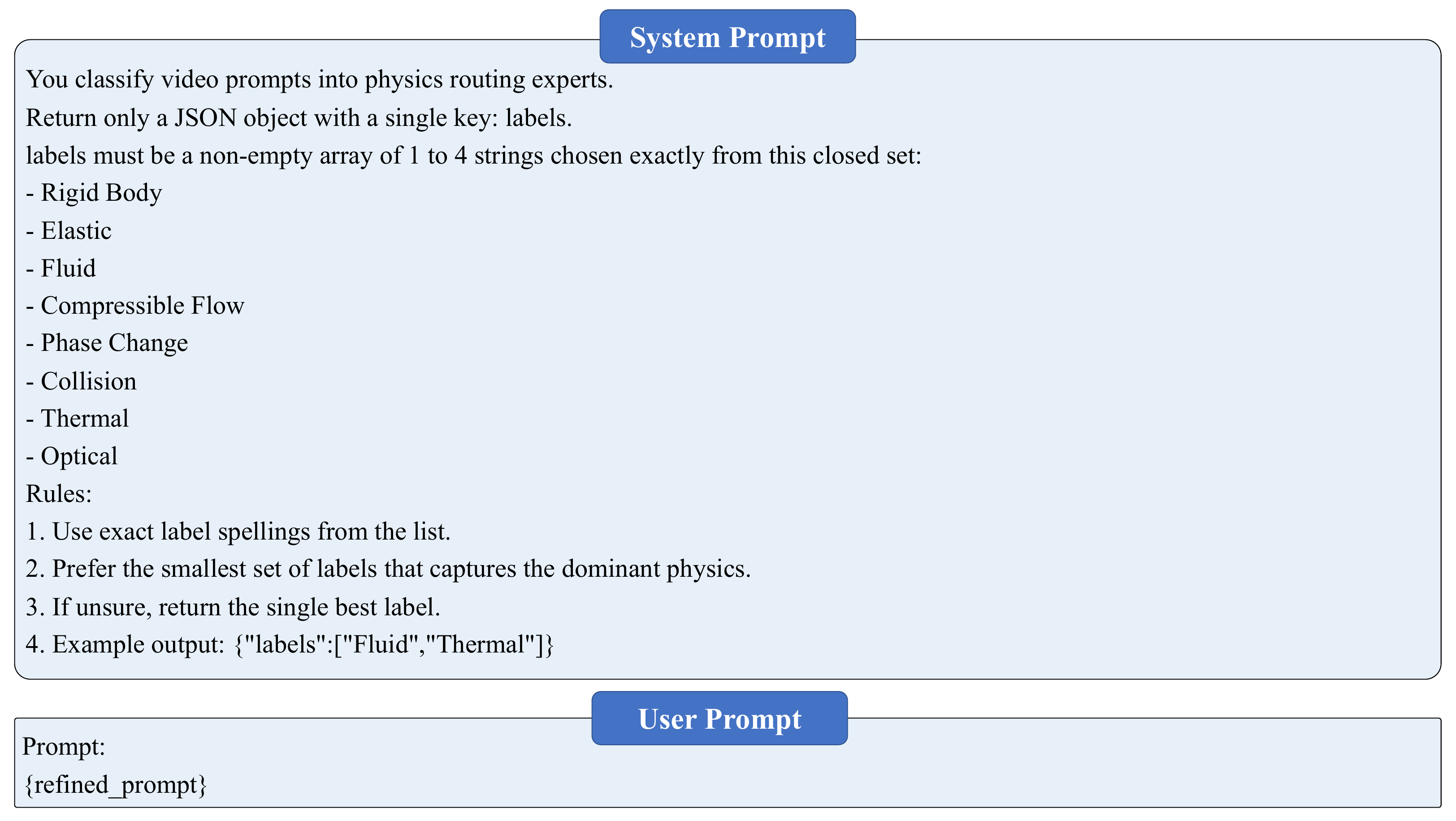}
    \caption{\textbf{LLM-based category-label estimation template for LPMER.} Given the refined physics-oriented description, the template asks the LLM to select the relevant physical categories from the eight-category expert set $\mathcal{P}$. The resulting multi-label set $\mathcal{Y}$ is used as label-prior evidence for routing.}
    \vspace{-0.3cm}
    \label{fig:lpmer_label_estimation}
\end{figure*}

Figures~\ref{fig:lpmer_prompt_refine} and~\ref{fig:lpmer_label_estimation} show the two prompt templates used by this inference-time parser. The first template rewrites the raw generation prompt into a physics-oriented description, and the second template maps that refined description to the active category set $\mathcal{Y}$ used by the LPMER label prior.

The label set is encoded as an additive routing prior rather than as a hard expert mask. Given the condition vector $\mathbf{c}$, the learned router predicts condition-dependent logits
\begin{equation}
    \mathbf{r}=R_{\psi}(\mathbf{c})\in\mathbb{R}^{|\mathcal{P}|}.
    \label{eq:appendix_lpmer_learned_logits}
\end{equation}
The multi-label prior adds a positive bias to the logits of active labels:
\begin{equation}
    b_{\mathrm{label},k}
    =
    \beta\,\mathbf{1}[k\in\mathcal{Y}],
    \qquad
    \boldsymbol{\ell}
    =
    \mathbf{r}+\mathbf{b}_{\mathrm{label}},
    \label{eq:appendix_lpmer_label_prior}
\end{equation}
where $\beta$ controls the strength of the label prior. This design lets prompt-derived labels guide expert selection while still allowing the learned router to assign capacity to complementary physical categories when their condition-dependent evidence is strong.

LPMER then performs sparse top-$k$ activation:
\begin{equation}
    \mathcal{K}
    =
    \mathrm{TopK}(\boldsymbol{\ell}),
    \qquad
    w_k
    =
    \frac{\exp(\ell_k/\tau)}
    {\sum_{m\in\mathcal{K}}\exp(\ell_m/\tau)},
    \quad k\in\mathcal{K},
    \label{eq:appendix_lpmer_topk}
\end{equation}
where $\tau$ is the routing temperature. In the default setting, $k=4$, which keeps the physics branch sparse while allowing multi-category scenes to activate more than one operator. This is important for coupled cases such as bouncing, pouring, boiling, or contact-driven deformation, where a single category label would be too restrictive.

Each selected operator receives the shared bank $\mathbf{a}$ and condition vector $\mathbf{c}$, and predicts a residual update restricted by its recipe mask:
\begin{equation}
    \begin{aligned}
    \Delta\mathbf{a}_k
    &=
    \mathbf{M}_k\odot O_k([\mathbf{c},\mathbf{a}]),
    \qquad
    \boldsymbol{\phi}_k
    =
    P_k(\mathbf{a}+\Delta\mathbf{a}_k),
    \\
    \tilde{\mathbf{a}}
    &=
    \mathbf{a}
    +
    \sum_{k\in\mathcal{K}}w_k\,\Delta\mathbf{a}_k .
    \end{aligned}
    \label{eq:appendix_lpmer_refined_bank}
\end{equation}
The mask $\mathbf{M}_k$ is defined by the same field recipe as the residual family of expert $k$, so routing changes which field-proxy slots can be updated and which residual constraints are eligible downstream. Here, $P_k$ is the category-specific field readout that forms the expert-specific field-proxy representation $\boldsymbol{\phi}_k$ used for residual evaluation. The routed operational residual loss is then evaluated on the label-consistent subset of selected experts, matching the main objective in Eq.~\ref{eq:physics_residual}. Thus, LPMER provides a soft, multi-label routing prior for operator selection, while CSRC supplies category-specific residuals on the corresponding expert-specific field-proxy representations.

\begin{table}[t]
    \centering
    \scriptsize
    \setlength{\tabcolsep}{2pt}
    \renewcommand{\arraystretch}{0.98}
    \caption{\textbf{Category-wise glossary of operational constraints.}}
    \label{tab:constraint_glossary}
    \begin{tabularx}{\linewidth}{@{} l X l X @{}}
    \toprule
    \textbf{Physical Category} & \textbf{Constraint} & \textbf{Name} & \textbf{Role} \\
    \midrule
    \multirow{4}{*}{Rigid Body}
    & $\partial_t d_k-u_k$ & Kinematic consistency & Aligns displacement-like and velocity-like slots. \\
    & $\varepsilon_k$ & Low-strain prior & Suppresses internal deformation. \\
    & $\sigma_k-\gamma_{\sigma}\varepsilon_k$ & Stress-strain closure & Keeps stress-like and strain-like proxies compatible up to latent scale. \\
    & $\mathrm{sym}\nabla u_k$ & Rigid-motion prior & Discourages locally deforming motion while allowing rotation-like components. \\
    \midrule
    \multirow{4}{*}{Elastic}
    & $\partial_t^2 d_k-c_d^2\Delta d_k$ & Displacement wave & Encourages coherent oscillatory deformation. \\
    & $\partial_t^2 u_k-c_u^2\Delta u_k$ & Velocity wave & Aligns motion with elastic propagation. \\
    & $\varepsilon_k-\mathrm{sym}\nabla d_k$ & Deformation compatibility & Couples strain-like and displacement-like proxies. \\
    & $\sigma_k-\gamma_{\sigma}\varepsilon_k$ & Stress-strain closure & Stabilizes elastic state compatibility up to latent scale. \\
    \midrule
    \multirow{4}{*}{Fluid}
    & $\partial_t \rho_k+\nabla\!\cdot(\rho_k u_k)$ & Mass continuity & Enforces transport-consistent density change. \\
    & $\partial_t u_k+(u_k\!\cdot\!\nabla)u_k+\nabla p_k/\rho_k-\nu\Delta u_k$ & Momentum balance & Biases velocity toward fluid-style evolution. \\
    & $\nabla p_k$ & Pressure regularity & Suppresses pressure spikes. \\
    & $\nabla \rho_k+\mathrm{ReLU}(\rho_0-\rho_k)$ & Density stability & Smooths density and discourages collapse. \\
    \midrule
    \multirow{4}{*}{Compressible Flow}
    & $\partial_t \rho_k+\nabla\!\cdot(\rho_k u_k)$ & Mass continuity & Tracks transported mass variation. \\
    & $\partial_t(\nabla\!\cdot u_k)-c_p\Delta p_k$ & Compression dynamics & Couples local compression changes to pressure-like variation. \\
    & $p_k-\gamma_p\rho_k$ & Pressure-density closure & Ties pressure-like and density-like slots together up to latent scale. \\
    & $\nabla u_k$ & Gradient regularity & Stabilizes expansion and compression patterns. \\
    \midrule
    \multirow{4}{*}{Phase Change}
    & $\partial_t\alpha_k+\nabla\!\cdot(\alpha_k u_k)$ & Phase transport & Couples support change to local motion. \\
    & $\partial_t T_k+u_k\!\cdot\nabla T_k-\kappa\Delta T_k-\beta\partial_t\alpha_k$ & Thermal-phase coupling & Links heat transport with support change. \\
    & $\rho_k-\gamma_{\rho\alpha}\alpha_k$ & Density-phase closure & Keeps density-like and phase/support slots aligned up to latent scale. \\
    & $\nabla\alpha_k$ & Support regularity & Stabilizes the phase/support proxy. \\
    \midrule
    \multirow{4}{*}{Collision}
    & $\partial_t d_k-u_k$ & Kinematic consistency & Aligns geometry-like and motion-like slots. \\
    & $\partial_t u_k-\gamma_j j_k$ & Impulse response & Couples motion change to impulse up to latent scale. \\
    & $\nabla\!\cdot d_k+\bar{j}_k$ & Contact compression & Encourages compression to match impact. \\
    & $\Delta j_k$ & Impulse propagation & Smooths impact transfer in space-time. \\
    \midrule
    \multirow{2}{*}{Thermal}
    & $\partial_t T_k+\nabla\!\cdot(u_k\,\bar{T}_k)-\kappa\Delta T_k$ & Heat transport & Couples thermal storage, advection, and diffusion. \\
    & $\nabla T_k$ & Thermal regularity & Discourages isolated temperature spikes. \\
    \midrule
    \multirow{3}{*}{Optical}
    & $\partial_t^2 \psi_k-c_{\psi}^2\Delta \psi_k$ & Wave propagation & Encourages coherent optical propagation. \\
    & $(1-\alpha_k)\psi_k$ & Support gating & Keeps wave-like activation inside the support proxy. \\
    & $\Delta \alpha_k$ & Support smoothness & Smooths the associated support field. \\
    \bottomrule
    \end{tabularx}
\end{table}

\subsection{Interpretation of Category-Specific Residual Constraints}
\label{sec:constraint_rationale}

We further clarify how to read the operational constraints in Table~\ref{tab:category_constraints}. In the current implementation, these expressions are evaluated on expert-specific field-proxy representations $\boldsymbol{\phi}_k$ rather than on externally supervised physical states. These residuals are not full solver-style governing equations. Instead, CSRC uses four operational constraint families: PDE-style anchors, kinematic consistency terms, closure proxies, and stabilizing priors. PDE-style anchors approximate balance or transport relations, including mass and momentum conservation, heat transport, and wave propagation. Kinematic consistency terms capture definitional links between slots, such as displacement--velocity relations. Closure proxies tie weakly observed slots together up to latent scale, while stabilizing priors suppress high-frequency or unsupported side channels. Table~\ref{tab:constraint_glossary} unpacks the main table term by term, while Table~\ref{tab:constraint_status} states the reference relation or prior from which each operational residual is abstracted.

\begin{table}[t]
    \centering
    \scriptsize
    \setlength{\tabcolsep}{2pt}
    \renewcommand{\arraystretch}{0.98}
    \caption{\textbf{Residual families and reference physical relations.}}
    \label{tab:constraint_status}
    \begin{tabularx}{\linewidth}{@{} l X l X @{}}
    \toprule
    \textbf{Physical Category} & \textbf{Constraint} & \textbf{Residual family} & \textbf{Reference Equation / Relation} \\
    \midrule
    \multirow{4}{*}{Rigid Body}
    & $\partial_t d_k-u_k$ & Kinematic consistency & $\partial_t d = u$ \\
    & $\varepsilon_k$ & Stabilizing prior & $\varepsilon\simeq0$ for rigid motion \\
    & $\sigma_k-\gamma_{\sigma}\varepsilon_k$ & Closure proxy & $\sigma\simeq C:\varepsilon$ with latent-scale stiffness \\
    & $\mathrm{sym}\nabla u_k$ & Stabilizing prior & $\mathrm{sym}\,\nabla u\simeq0$ \\
    \midrule
    \multirow{4}{*}{Elastic}
    & $\partial_t^2 d_k-c_d^2\Delta d_k$ & PDE-style anchor & $\partial_t^2 d-c_d^2\Delta d=0$ \\
    & $\partial_t^2 u_k-c_u^2\Delta u_k$ & PDE-style anchor & $\partial_t^2 u-c_u^2\Delta u=0$ \\
    & $\varepsilon_k-\mathrm{sym}\nabla d_k$ & Kinematic consistency & $\varepsilon=\mathrm{sym}\,\nabla d$ \\
    & $\sigma_k-\gamma_{\sigma}\varepsilon_k$ & Closure proxy & $\sigma\simeq C:\varepsilon$ with latent-scale stiffness \\
    \midrule
    \multirow{4}{*}{Fluid}
    & $\partial_t \rho_k+\nabla\!\cdot(\rho_k u_k)$ & PDE-style anchor & $\partial_t\rho+\nabla\!\cdot(\rho u)=0$ \\
    & $\partial_t u_k+(u_k\!\cdot\!\nabla)u_k+\nabla p_k/\rho_k-\nu\Delta u_k$ & PDE-style anchor & $\partial_t u+(u\!\cdot\!\nabla)u+\nabla p/\rho-\nu\Delta u=0$ \\
    & $\nabla p_k$ & Stabilizing prior & bounded pressure-gradient proxy \\
    & $\nabla \rho_k+\mathrm{ReLU}(\rho_0-\rho_k)$ & Stabilizing prior & $\rho>0$ and locally smooth density \\
    \midrule
    \multirow{4}{*}{Compressible Flow}
    & $\partial_t \rho_k+\nabla\!\cdot(\rho_k u_k)$ & PDE-style anchor & $\partial_t\rho+\nabla\!\cdot(\rho u)=0$ \\
    & $\partial_t(\nabla\!\cdot u_k)-c_p\Delta p_k$ & PDE-style anchor & $\partial_t(\nabla\!\cdot u)-c_p\Delta p\simeq0$ \\
    & $p_k-\gamma_p\rho_k$ & Closure proxy & $p\simeq c^2\rho$ up to latent scaling \\
    & $\nabla u_k$ & Stabilizing prior & bounded compression gradients \\
    \midrule
    \multirow{4}{*}{Phase Change}
    & $\partial_t\alpha_k+\nabla\!\cdot(\alpha_k u_k)$ & PDE-style anchor & $\partial_t\alpha+\nabla\!\cdot(\alpha u)=S_{\alpha}$ with unresolved source \\
    & $\partial_t T_k+u_k\!\cdot\nabla T_k-\kappa\Delta T_k-\beta\partial_t\alpha_k$ & PDE-style anchor & heat transport with latent phase-change coupling \\
    & $\rho_k-\gamma_{\rho\alpha}\alpha_k$ & Closure proxy & $\rho\simeq\rho(\alpha)$ up to latent scaling \\
    & $\nabla\alpha_k$ & Stabilizing prior & smooth phase/support proxy \\
    \midrule
    \multirow{4}{*}{Collision}
    & $\partial_t d_k-u_k$ & Kinematic consistency & $\partial_t d = u$ \\
    & $\partial_t u_k-\gamma_j\,j_k$ & Closure proxy & $m\,\partial_t u\simeq j$ after absorbing scale \\
    & $\nabla\!\cdot d_k+\bar{j}_k$ & Closure proxy & $\nabla\!\cdot d+\bar{j}\simeq0$ \\
    & $\Delta j_k$ & Stabilizing prior & spatial impulse smoothness \\
    \midrule
    \multirow{2}{*}{Thermal}
    & $\partial_t T_k+\nabla\!\cdot(u_k\,\bar{T}_k)-\kappa\Delta T_k$ & PDE-style anchor & $\partial_t T+\nabla\!\cdot(uT)-\kappa\Delta T=Q$ with unresolved source \\
    & $\nabla T_k$ & Stabilizing prior & bounded temperature-gradient proxy \\
    \midrule
    \multirow{3}{*}{Optical}
    & $\partial_t^2 \psi_k-c_{\psi}^2\Delta \psi_k$ & PDE-style anchor & $\partial_t^2\psi-c^2\Delta\psi=0$ \\
    & $(1-\alpha_k)\psi_k$ & Stabilizing prior & wave proxy should remain inside active support \\
    & $\Delta \alpha_k$ & Stabilizing prior & smooth optical support \\
    \bottomrule
    \end{tabularx}
\end{table}

\paragraph{1. Rigid Body.}
The rigid-body expert is derived from rigid kinematics rather than from a deformable material model. The residual $\partial_t d_k-u_k$ comes directly from the displacement--velocity relation $\partial_t d=u$. A rigid body should have negligible internal strain, so $\varepsilon_k$ acts as a low-strain prior and $\sigma_k-\gamma_{\sigma}\varepsilon_k$ keeps the stress-like proxy compatible with that regime up to latent scale. The term $\mathrm{sym}\nabla u_k$ is a finite-difference stabilizing prior for local non-rigidity; using the symmetric part avoids penalizing rotation-like components as strongly as a full velocity-gradient penalty.

\paragraph{2. Elastic.}
The elastic expert abstracts the small-deformation wave behavior implied by linear elasticity. In a homogeneous medium, the displacement field satisfies a wave-like equation after simplifying the elastodynamic balance; this motivates $\partial_t^2 d_k-c_d^2\Delta d_k$, and the same propagation bias is applied to the velocity slot through $\partial_t^2 u_k-c_u^2\Delta u_k$. The term $\varepsilon_k-\mathrm{sym}\nabla d_k$ is an operational compatibility proxy for $\varepsilon=\mathrm{sym}\,\nabla d$, adapted to slot-valued latent fields. The stress-strain term $\sigma_k-\gamma_{\sigma}\varepsilon_k$ implements a Hooke-style closure $\sigma\simeq C:\varepsilon$ without estimating a full stiffness tensor.

\paragraph{3. Fluid.}
The fluid expert follows the standard mass and momentum balances used in viscous flow. The continuity residual $\partial_t\rho_k+\nabla\!\cdot(\rho_k u_k)$ is the conservative form of mass conservation. The momentum residual is a Navier--Stokes-style balance containing temporal acceleration, convective transport, pressure forcing, and viscous diffusion. Since pressure and density are constructed proxies, $\nabla p_k$ suppresses pressure spikes and $\nabla\rho_k+\mathrm{ReLU}(\rho_0-\rho_k)$ encourages smooth positive density. These stabilizers are not additional governing laws; they keep the latent fields numerically usable for residual evaluation.

\paragraph{4. Compressible Flow.}
The compressible-flow expert keeps mass conservation but replaces the incompressible-style momentum emphasis with terms that expose expansion and compression. The continuity term again comes from $\partial_t\rho+\nabla\!\cdot(\rho u)=0$. The residual $\partial_t(\nabla\!\cdot u_k)-c_p\Delta p_k$ tracks temporal changes of local compression and couples them to pressure-like variation under an acoustic or barotropic linearization. The proxy $p_k-\gamma_p\rho_k$ is a lightweight equation-of-state surrogate, corresponding to $p\simeq c^2\rho$ up to latent scaling. The gradient term $\nabla u_k$ prevents unstable compression patterns from becoming a high-frequency side channel.

\paragraph{5. Phase Change.}
The phase-change expert is motivated by transported phase indicators and Stefan-style thermal coupling. The phase slot $\alpha_k$ is treated as an occupancy/support proxy, so $\partial_t\alpha_k+\nabla\!\cdot(\alpha_k u_k)$ links support change to local motion, standing in for an advective phase-balance equation with unresolved source terms. The combined thermal term $\partial_t T_k+u_k\!\cdot\nabla T_k-\kappa\Delta T_k-\beta\partial_t\alpha_k$ couples storage, advection, diffusion, and latent phase-change activity without claiming calibrated heat capacity or latent heat. The proxy $\rho_k-\gamma_{\rho\alpha}\alpha_k$ ties density to phase occupancy, reflecting that different phases induce different density/support patterns.

\paragraph{6. Collision.}
The collision expert encodes impulse-response structure rather than solving complementarity contact conditions. The kinematic term $\partial_t d_k-u_k$ again enforces the relation between displacement and velocity. The impulse term $\partial_t u_k-\gamma_j\,j_k$ is derived from the impulse-momentum relation $m\,\partial_t u\simeq j$ after absorbing mass and scale into the latent proxy. The compression term $\nabla\!\cdot d_k+\bar{j}_k$ encourages converging displacement patterns to coincide with impact activity, while $\Delta j_k$ smooths the impulse field so that contact response propagates locally instead of appearing as isolated noise.

\paragraph{7. Thermal.}
The thermal expert uses a compact heat-transport proxy. The term $\partial_t T_k+\nabla\!\cdot(u_k\,\bar{T}_k)-\kappa\Delta T_k$ combines temporal storage, conservative advection, and diffusion, while $\nabla T_k$ limits isolated temperature spikes. We omit unknown heat sources, material heat capacity, and boundary fluxes, so the resulting residual is a low-frequency thermal prior rather than a complete energy-balance solve.

\paragraph{8. Optical.}
The optical expert abstracts wave-like propagation into a scalar latent proxy. The residual $\partial_t^2\psi_k-c_{\psi}^2\Delta\psi_k$ follows the scalar wave equation obtained after suppressing constants and polarization details. The support-gating term $(1-\alpha_k)\psi_k$ discourages wave-like activation outside the associated support proxy, while $\Delta\alpha_k$ smooths that support. These terms encourage coherent propagation and support consistency without modeling full Maxwell dynamics.

\paragraph{Unified interpretation.}
Across all eight physical categories, these terms act as structured inductive bias rather than as certificates of exact physical correctness. They are evaluated by fixed finite-difference operators on expert-specific field-proxy representations, remain tolerant to pseudo-motion noise and missing boundary conditions, and give the frozen generator an interpretable physical interface on which constraint alignment can operate. In that sense, the residual families serve as a practical bridge between constructed field proxies and pretrained generative dynamics.

\section{Training Details}

\subsection{Training Dataset and Physical-Category Relabeling}
\label{sec:dataset_taxonomy}

Our training data is based on WISA-80K~\cite{wang2025wisa}. We do not merge additional video datasets into the training split. Instead, we reorganize the WISA-80K clips under eight physical categories used by LPMER and CSRC. These categories consist of four material categories (Rigid Body, Elastic, Fluid, and Compressible Flow) and four phenomenon categories (Collision, Phase Change, Thermal, and Optical). This relabeling makes the training annotations consistent with the routed expert taxonomy in Table~\ref{tab:category_constraints}, while preserving the original video-text distribution of WISA-80K. The resulting category distribution is summarized in Fig.~\ref{fig:data}.

\begin{table}[t]
\centering
\small
\setlength{\tabcolsep}{3pt}
\renewcommand{\arraystretch}{1.05}
\caption{\textbf{Eight physical categories used to relabel WISA-80K for training.} The taxonomy contains four material categories and four phenomenon categories.}
\label{tab:wisa80k_relabeling}
\begin{tabularx}{\linewidth}{@{} l l X X @{}}
\toprule
\textbf{Group} & \textbf{Physical category} & \textbf{Physical criterion} & \textbf{Typical visual cues} \\
\midrule
Material & Rigid Body & Coherent translation/rotation with negligible internal deformation. & Falling, rolling, sliding, or spinning rigid objects. \\
Material & Elastic & Reversible deformation with strain and stress-like response. & Bouncing, stretching, bending, oscillation. \\
Material & Fluid & Liquid-like flow dominated by advection and pressure-driven motion. & Pouring, splashing, dripping, viscous flow. \\
Material & Compressible Flow & Density-varying gas, smoke, jet, or expansion/compression motion. & Smoke plumes, air jets, explosions, fire-like expansion. \\
Phenomenon & Collision & Multi-object interaction dominated by impact or contact impulse. & Bouncing contact, stacking, grasping, frictional sliding. \\
Phenomenon & Phase Change & State transition coupled to thermal or occupancy change. & Melting, freezing, boiling, vaporization, solidification. \\
Phenomenon & Thermal & Heat transfer or diffusion without a dominant phase transition. & Heating, cooling, thermal convection, glowing heat spread. \\
Phenomenon & Optical & Light, reflection, refraction, wave, or support-dependent optical effects. & Reflections, lenses, rainbows, caustics, interference-like patterns. \\
\bottomrule
\end{tabularx}
\end{table}

\noindent\textbf{Relabeling protocol.}
We assign physical-category labels from the video prompt, accompanying physical description, and observable motion pattern. Each clip receives the dominant physical category, and clips containing coupled mechanisms may retain multiple active labels so that LPMER can route several experts. For example, a bouncing ball can activate both Rigid Body and Collision, while boiling water can activate Phase Change, Fluid, and Thermal. The relabeling is therefore mechanism-oriented rather than surface-category-oriented.

\noindent\textbf{Use in training.}
The relabeled WISA-80K physical-category annotations provide the label priors used by the router and determine which residual families are eligible during physics-coupled training. They do not provide direct supervision for pressure, density, temperature, phase, impulse, or optical fields. Those fields remain operational proxies constructed through AFE and constrained by the residual terms described in Sec.~\ref{sec:constraint_rationale}.

\subsection{Implementation Details}
\label{sec:appendix_implementation_details}

\noindent\textbf{Backbones and data.}
We train the PILA adapter on Wan 2.1-T2V-1.3B~\cite{wan2025wan}. The pretrained video generator is kept frozen, and the training scripts optimize only the PILA adapter. Training clips are read from WISA-80K~\cite{wang2025wisa} with the relabeled eight-category annotations described in Sec.~\ref{sec:dataset_taxonomy}. Unless otherwise specified, training videos are resized to $480\times832$ and sampled as 81-frame clips. For Wan 2.2-T2V-A14B evaluation, we do not optimize a new adapter; we load the Wan 2.1-trained adapter and use it as a frozen correction module on the frozen Wan 2.2 backbone.

\noindent\textbf{Wan 2.1-1.3B training.}
For the 1.3B backbone, we use a staged schedule. The observable pretraining stage is run over progressively wider diffusion timestep ranges: $[0.90,1.00]$, $[0.75,1.00]$, and $[0.50,1.00]$. Each phase runs for 3K steps with learning rate $1\times10^{-5}$. The full physics-coupling stage is then trained over $[0.00,1.00]$ for 3 epochs with the same learning rate. We set the physics loss weight to $0.30$, use 2K physics warm-up steps and 1K conditioned-physics warm-up steps, freeze the physical encoder for the first 1K full-coupling steps, and then scale its learning rate by $0.3$. The default adapter hidden width is 128, the operational physical attribute bank has 32 channels, and the router selects the top-$4$ experts.

\noindent\textbf{Wan 2.2-14B transfer.}
For the 14B backbone, no Wan 2.2-specific adapter training is performed. We directly load the adapter trained on Wan 2.1-1.3B and keep both the Wan 2.2 generator and the adapter weights frozen during evaluation. The transferred adapter keeps the same 32-channel attribute bank, top-$4$ expert routing, finite-difference residual interface, sigma conditioning, and quadratic sigma gate with floor $0.05$. For the Wan 2.2 dual-expert scheduler, the high-/low-noise boundary is set to $0.417$ in scheduler-index space when applying the correction across denoising regimes.

\noindent\textbf{Optimization and residual evaluation.}
The training objective used for the Wan 2.1-1.3B adapter combines flow-matching supervision and routed operational residuals, following the loss in Eq.~\ref{eq:total_loss}. We optimize trainable adapter parameters using AdamW~\cite{loshchilov2017decoupled} with a constant learning-rate schedule, \texttt{bf16} mixed precision, and DeepSpeed ZeRO-2~\cite{rajbhandari2020zero} with CPU offload. Physics residuals are evaluated by fixed finite-difference operators on expert-specific field-proxy representations rather than by differentiating through a full external simulator. For Wan 2.2-14B, no optimizer steps are taken; the transferred adapter is evaluated directly as an inference-time correction module. During inference, the adapter predicts a lightweight correction to the frozen backbone flow-matching vector field, so the base generator weights remain unchanged.

\noindent\textbf{Compute resources.}
The Wan 2.1-1.3B adapter is trained on a single 8-GPU node with NVIDIA A100 GPUs, each with 80GB memory. Full WISA-80K training under the staged schedule takes approximately 40 hours. Unless otherwise stated, the 1.3B ablations use the same hardware configuration and training budget as the full adapter. The Wan 2.2-14B setting is evaluated by direct adapter transfer and does not require additional 14B-specific training.

\subsection{Training Procedure}
\label{sec:training_procedure}

We adopt a staged optimization strategy that separates observable physical encoding, unobserved field-proxy completion, and full flow-matching coupling. This staged training is used for the Wan 2.1-1.3B adapter, which is then transferred to Wan 2.2-14B without additional optimization. This design is critical for preserving the pretrained generator's visual quality while preventing hidden field-proxy slots from becoming unconstrained side channels.

\textbf{Stage A (Observable Attribute-Bank Pretraining).}
We first train the shared physical encoder, operational 32-channel attribute-bank head, and observable flow head using frozen-backbone latent estimates and dense flow/deformation proxy targets. The current Wan2.1 training script runs this observable pretraining across progressively wider noise ranges, so the bank learns stable flow-like slots before it is asked to support routed operational residuals. 

\textbf{Stage B (Encoder Completion and Field Recovery).}
After observable pretraining, we complete the field-proxy interface without updating the generator. AFE follows the field construction described in Sec.~\ref{sec:afe_field_construction}: the motion-derived velocity proxy provides the main kinematic anchor, pressure and density are constructed from kinematic descriptors, deformation-related proxies are derived from motion, and weakly observed slots are activated through the ordered phases $\texttt{core}\rightarrow\alpha\rightarrow T\rightarrow j\rightarrow\psi$. The recovery loss combines local only-u fluid terms, kinematic consistency, phase/thermal transport, impulse fitting, wave-propagation anchors, and a small observable anchor. The physical mask used in this stage is built from active routed fields and blended with the bootstrap motion mask.

\textbf{Stage C (Physics-to-FM Coupling).}
Finally, the completed encoder initializes the full PILA adapter. The pretrained FM backbone remains frozen; trainable components include the label/sigma condition encoder, top-$k$ router, masked operator experts, shared attribute decoder, and the final correction decoder. The router selects physical categories with label-prior bias, each selected expert emits a recipe-masked attribute update, CSRC regularizes the corresponding expert-specific field-proxy representations, and the refined physical attribute bank is decoded together with $v_{\theta_0}$ into $\Delta v_{\psi}$. The full-stage objective combines FM supervision and routed operational residuals on label-consistent expert-specific field-proxy representations.

\textbf{Practical considerations.}
In implementation, we apply operational residual constraints in the latent spatiotemporal resolution used by the Wan VAE rather than at pixel resolution, reducing sensitivity to noisy pseudo-motion estimates and preventing over-constraining appearance-level dynamics. The bootstrap motion mask is constructed as a blend of velocity time derivatives (55\% weight), latent-space derivatives (25\% weight), and spatial velocity gradients (20\% weight), then normalized by a per-sample quantile with a floor of 0.08. During full coupling, this mask is blended over the warmup with the active physical mask derived from routed fields. The MoE router uses top-$k=4$ selection with softmax normalization and label-prior bias. In the main Wan2.1 configuration, the operational bank has 32 channels, the adapter hidden width is 128, the physics weight target is 0.30, the physics warmup is 2000 steps, the conditioning warmup is 1000 steps, and the motion-mask warmup is 300 steps. The overall design provides a stable optimization path that preserves pretrained quality while injecting interpretable physics-aware guidance.

\begin{table}[t]
    \centering
    \caption{\textbf{AFE field-construction ablations on Wan 2.1-1.3B.} We ablate the kinematic anchor and staged recovery used to construct the operational physical attribute bank.}
    \setlength{\tabcolsep}{1pt}
    \renewcommand{\arraystretch}{1.0}
    \begin{tabular*}{\linewidth}{@{\extracolsep{\fill}} l | c c c c c | c | c @{}}
    \toprule
    \textbf{AFE Variant} & \textbf{Mechanics}↑ & \textbf{Optics}↑ & \textbf{Thermal}↑ & \textbf{Material}↑ & \textbf{Average}↑ & \textbf{Imaging}↑ & \textbf{Motion}↑\\
    \midrule
    Baseline & 0.483 & 0.640 & 0.322 & 0.342 & 0.467 & 0.538 & 0.250  \\
    w/o Kinematic Anchor & 0.518 & \underline{0.647} & 0.553 & 0.527 & \underline{0.567} & \underline{0.542}  & 0.324 \\
    w/o Staged Recovery & \underline{0.523} & 0.623 & 0.545 & \underline{0.531} & 0.560 & 0.541  & \underline{0.331} \\
    PILA (Full AFE) & \textbf{0.575} & \textbf{0.713} & \textbf{0.578} & \textbf{0.558} & \textbf{0.615} & \textbf{0.543}  & \textbf{0.356}  \\
    \bottomrule
    \end{tabular*}
    \vspace{-0.4cm}
    \label{tab:ablation_encoder}
\end{table}

\begin{table}[t]
    \centering
    \small
    \caption{\textbf{Operational residual-family ablations on Wan 2.1-1.3B.} Each row removes one residual subfamily, using the grouping clarified in Tables~\ref{tab:constraint_glossary}--\ref{tab:constraint_status}, while keeping the same backbone and evaluation protocol.}
    \setlength{\tabcolsep}{1pt}
    \renewcommand{\arraystretch}{1.0}
    \begin{tabular*}{\linewidth}{@{\extracolsep{\fill}} l | c c c c c | c | c @{}}
    \toprule
    \textbf{Configuration} & \textbf{Mechanics}↑ & \textbf{Optics}↑ & \textbf{Thermal}↑ & \textbf{Material}↑ & \textbf{Average}↑ & \textbf{Imaging}↑ & \textbf{Motion}↑\\
    \midrule
    Baseline & 0.483 & 0.640 & 0.322 & 0.342 & 0.467 & 0.538 & 0.250  \\
    Proxy-only fields & 0.518 & 0.647 & 0.543 & 0.527 & 0.565 & 0.541  & 0.306 \\
    w/o PDE-style anchors & 0.539 & 0.679 & 0.550 & \underline{0.544} & 0.586 &  0.539 & 0.345 \\
    w/o closure proxies & 0.552 & 0.691 & 0.546 & 0.543 & 0.592 & 0.535  & 0.353 \\
    w/o kinematic consistency & 0.547 & \underline{0.696} & 0.552 & 0.548 & \underline{0.595} &  \underline{0.542} & 0.349 \\
    w/o stabilizing priors & \underline{0.563} & 0.665 & \underline{0.576} & 0.542 & 0.593 &  0.540 & \underline{0.354} \\
    Full Model & \textbf{0.575} & \textbf{0.713} & \textbf{0.578} & \textbf{0.558} & \textbf{0.615} & \textbf{0.543}  & \textbf{0.356}  \\
    \bottomrule
    \end{tabular*}
    \vspace{-0.2cm}
    \label{tab:ablation_residual}
\end{table}

\begin{table}[t]
    \centering
    \caption{\textbf{Additional routing-label ablations on Wan 2.1-1.3B.} All variants use the same backbone and evaluation protocol.}
    \setlength{\tabcolsep}{1pt}
    \renewcommand{\arraystretch}{1.0}
    \begin{tabular*}{\linewidth}{@{\extracolsep{\fill}} l |c c c c c | c | c @{}}
    \toprule
    \textbf{Configuration} & \textbf{Mechanics}↑ & \textbf{Optics}↑ & \textbf{Thermal}↑ & \textbf{Material}↑ & \textbf{Average}↑ & \textbf{Imaging}↑ & \textbf{Motion}↑\\
    \midrule
    Baseline & 0.483 & 0.640 & 0.322 & 0.342 & 0.467 & \underline{0.538} & 0.250  \\
    w/ Random Label & \underline{0.500} & \underline{0.687} & \underline{0.570} & \underline{0.542} & \underline{0.582} & 0.532  & \underline{0.307}  \\
    w/ LLM label & \textbf{0.575} & \textbf{0.713} & \textbf{0.578} & \textbf{0.558} & \textbf{0.615} & \textbf{0.543} & \textbf{0.356}\\
    \bottomrule
    \end{tabular*}
    \vspace{-0.2cm}
    \label{tab:ablation_router}
\end{table}

\section{Additional Experiment Details and Analysis}

\subsection{Supplementary Ablation Analysis}
\label{sec:appendix_ablation_details}

This section expands the main ablation study into module-level controls. All variants use the Wan 2.1-1.3B backbone and the same evaluation protocol. Table~\ref{tab:ablation} reports the main component ablations, while Tables~\ref{tab:ablation_encoder}--\ref{tab:ablation_router} give the focused analyses for the three core modules: AFE field construction, CSRC residual supervision, and LPMER label-prior routing. The goal is to isolate where the physical improvements come from: added adapter capacity, the operational physical interface, category-specific operator refinement, residual supervision, routing-label quality, and the final physics-to-flow injection.

\begin{table}[t]
    \centering
    \small
    \setlength{\tabcolsep}{3pt}
    \renewcommand{\arraystretch}{1.05}
    \caption{\textbf{Definitions of ablation variants.}}
    \label{tab:appendix_ablation_definitions}
    \begin{tabularx}{\linewidth}{@{} l X X @{}}
    \toprule
    \textbf{Variant} & \textbf{Changed component} & \textbf{Diagnostic purpose} \\
    \midrule
    Baseline & Uses the frozen Wan 2.1-1.3B generator without the PILA adapter. & Measures the pretrained backbone before adding any trainable physics pathway. \\
    + Plain Adapter & Keeps an extra trainable adapter but removes the structured physical bank, routed operators, and residual constraints. & Tests whether gains come mainly from added capacity. \\
    w/o Flow correction & Removes the final decoder that injects the refined physical attribute bank into the FM vector field. & Tests whether the constructed operational physical interface actually affects the generated trajectory. \\
    w/o AFE & Removes the dedicated physical encoder pathway before AFE. & Tests whether a physical feature extractor is needed beyond a generic adapter branch. \\
    w/o MoE operator & Removes category-specific masked operator refinement. & Tests whether routed operator specialization matters beyond a shared physical bank. \\
    w/o Residual Constraints & Removes CSRC residual supervision during training. & Tests whether operational residuals shape the adapter beyond FM supervision alone. \\
    w/o Prompt Refine & Uses the routing-label pipeline without prompt refinement. & Tests sensitivity to cleaner prompt-derived physical-category evidence. \\
    w/o Expert Routing & Disables sparse top-$k$ expert selection and uses a non-routed expert path. & Tests whether label-prior routing is useful after the physical bank is present. \\
    w/o Kinematic Anchor & Removes the motion-derived velocity proxy used as the primary AFE construction anchor. & Tests whether AFE needs an observable kinematic anchor before constructing derived field proxies. \\
    w/o Staged Recovery & Uses a non-staged field recovery path rather than the observable-first AFE recovery schedule. & Tests whether progressively activating weakly observed slots is needed for stable field construction. \\
    w/ Random Label & Replaces physical-category labels with random labels for routing. & Tests whether the router gains come from meaningful physical labels rather than label noise. \\
    w/ LLM label & Uses LLM-inferred physical-category labels for routing. & Tests whether prompt-level physical semantics provide useful routing priors. \\
    w/o Prompt Refine & Removes the prompt-refinement step while retaining the routing-label interface. & Tests whether refining prompt evidence affects routing robustness. \\
    Full Model & Uses AFE, LPMER, masked operators, CSRC, and physics-to-flow correction. & Reports the complete PILA configuration used in the main comparison. \\
    \bottomrule
    \end{tabularx}
    \vspace{-0.3cm}
\end{table}

\noindent\textbf{Explanation of key component ablations.}
Table~\ref{tab:appendix_ablation_definitions} defines each variant used in Table~\ref{tab:ablation}. The plain adapter control separates PILA from a capacity-only explanation. The physical-encoder and flow-correction controls test the two ends of the physical pathway: whether a dedicated physical feature is constructed, and whether the refined physical attribute bank is actually injected back into the FM vector field. The MoE-operator, residual-family, and expert-routing controls then isolate the middle of the pipeline, where category-specific updates and residual losses are applied to the shared bank.

\noindent\textbf{Ablation of anchored field estimation (AFE) strategy.}
Table~\ref{tab:ablation_encoder} expands the AFE ablation beyond the single main-table component removal. It asks whether the operational physical bank should be built from an observable kinematic anchor and whether weakly observed fields should be introduced through staged recovery. Removing the kinematic anchor still improves over the frozen baseline because the adapter can learn generic corrections, but it lowers the average physical score from $0.615$ to $0.567$ and the motion score from $0.356$ to $0.324$. Removing staged recovery similarly reduces the average score to $0.560$. These drops indicate that AFE is not simply a free 32-channel side branch: the motion-derived proxy grounds the downstream field constructors, and staged recovery prevents pressure, density, thermal, phase, impulse, and wave slots from becoming unconstrained latent channels. Imaging quality remains nearly unchanged across these variants, so the AFE gain mainly comes from better physical organization rather than a visual-quality trade-off.

\noindent\textbf{Ablation of label-prior masked expert routing (LPMER) strategy.}
Table~\ref{tab:ablation_router} focuses on the label-prior side of LPMER. Random labels intentionally break the semantic correspondence between prompts and routed experts. This setting still improves over the frozen baseline because the adapter and expert pathway remain active, but it trails the LLM-label setting in both average physical score ($0.582$ vs. $0.615$) and motion score ($0.307$ vs. $0.356$). The gap shows that LPMER is not only adding sparse expert capacity; it benefits from physically meaningful prompt-derived category evidence. This table complements the ``w/o Expert Routing'' and ``w/o Prompt Refine'' rows in Table~\ref{tab:ablation}: those rows test whether sparse routing and prompt refinement are useful, while Table~\ref{tab:ablation_router} isolates whether the routed labels themselves carry useful physical semantics.

\noindent\textbf{Ablation of category-specific residual constraints (CSRC) strategy.}
Table~\ref{tab:ablation_residual} decomposes CSRC into the residual families used by the routed experts. The proxy-only variant keeps the constructed field proxies but removes the structured residual-family decomposition; it improves over the no-adapter baseline but remains far below the full model ($0.565$ vs. $0.615$ average), showing that proxy fields alone are insufficient. Removing PDE-style anchors, closure proxies, kinematic consistency terms, or stabilizing priors each reduces the average score relative to the full CSRC objective. The effect is distributed rather than dominated by one term family: PDE-style anchors support transport, diffusion, momentum, heat, and wave behavior; closure proxies couple stress, strain, pressure, density, phase, and impulse proxies up to latent scale; kinematic consistency keeps motion and displacement-like fields coherent; and stabilizing priors suppress noisy finite-difference artifacts and unsupported side channels. The full CSRC combination gives the best score on every physics metric while preserving imaging quality, supporting the claim that the residuals act as complementary constraints on the same operational physical interface.

\begin{figure}[t]
    \centering
    \includegraphics[width=\linewidth]{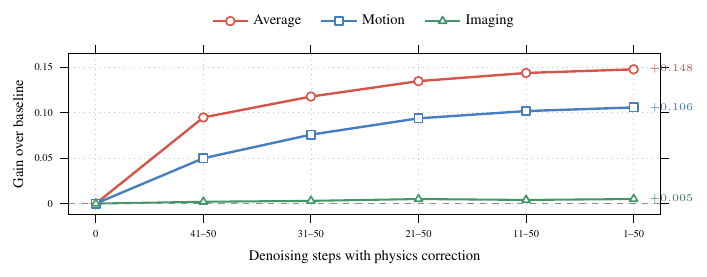}
    \caption{\textbf{Effect of the denoising-step schedule.} Gains are computed relative to the no-adapter baseline in Table~\ref{tab:ablation_denoise}. Expanding the sigma-gated correction window improves physical average and motion scores while preserving imaging quality.}
    \label{fig:diffusion_step}
\end{figure}

\noindent\textbf{Denoising-step schedule.}
Figure~\ref{fig:diffusion_step} visualizes the denoising-step sensitivity reported in Table~\ref{tab:ablation_denoise}. As the correction range expands from late-only steps to the full 1--50 schedule, physics-oriented average and motion scores improve steadily, while imaging quality remains nearly unchanged. This trend supports the sigma-conditioned and sigma-gated correction strategy: early noisy states are handled conservatively, but allowing the adapter to operate throughout denoising lets physical alignment accumulate across the sampling trajectory.

\begin{table}[t]
    \centering
    \caption{\textbf{Denoising-step sensitivity of the physics adapter on Wan 2.1-1.3B.} We vary the denoising-step range in which sigma-gated physical corrections are applied while keeping the same backbone and evaluation protocol.}
    \setlength{\tabcolsep}{1pt}
    \renewcommand{\arraystretch}{1.0}
    \begin{tabular*}{\linewidth}{@{\extracolsep{\fill}} l |c c c c c | c | c @{}}
    \toprule
    \textbf{Adding Steps} & \textbf{Mechanics}↑ & \textbf{Optics}↑ & \textbf{Thermal}↑ & \textbf{Material}↑ & \textbf{Average}↑ & \textbf{Imaging}↑ & \textbf{Motion}↑\\
    \midrule
    0     & 0.483 & 0.640 & 0.322 & 0.342 & 0.467 & 0.538 & 0.250 \\
    41-50 & 0.535 & 0.673 & 0.500 & 0.492 & 0.562 & 0.540 & 0.300 \\
    31-50 & 0.558 & 0.695 & 0.540 & 0.520 & 0.585 & 0.541 & 0.326 \\
    21-50 & 0.567 & 0.704 & 0.560 & 0.540 & 0.602 & 0.543 & 0.344 \\
    11-50 & \underline{0.572} & \underline{0.710} & \underline{0.570} & \underline{0.552} & \underline{0.611} & \underline{0.542} & \underline{0.352} \\
    1-50  & \textbf{0.575} & \textbf{0.713} & \textbf{0.578} & \textbf{0.558} & \textbf{0.615} & \textbf{0.543} & \textbf{0.356} \\
    \bottomrule
    \end{tabular*}
    \vspace{-0.2cm}
    \label{tab:ablation_denoise}
\end{table}

\noindent\textbf{Top-$k$ routing sensitivity.}
Table~\ref{tab:appendix_topk_fig} gives the numeric values behind Fig.~\ref{fig:expert_tradeoff}. Increasing $k$ activates more physical-category experts and gradually improves the average physics-oriented score, but the gain becomes small after $k=4$. The marginal increase from $k=4$ to $k=8$ is modest relative to the additional latency, so the main experiments use $k=4$ as the default quality-efficiency balance.

\begin{table}[!ht]
    \centering
    \caption{\textbf{Sensitivity to top-$k$ expert selection.}}
    \setlength{\tabcolsep}{4pt}
    \renewcommand{\arraystretch}{1.0}
    \begin{tabular*}{\linewidth}{@{\extracolsep{\fill}} l c c c c c c c c c @{}}
    \toprule
    \textbf{$k$} & \textbf{0} & \textbf{1} & \textbf{2} & \textbf{3} & \textbf{4} & \textbf{5} & \textbf{6} & \textbf{7} & \textbf{8} \\
    \midrule
    \textbf{Average}↑ & 0.592 & 0.602 & 0.610 & 0.612 & 0.615 & 0.617 & 0.621 & 0.621 & 0.623 \\
    \textbf{Time (s)} & 192.63 & 193.80 & 194.79 & 195.83 & 196.03 & 196.20 & 196.95 & 197.84 & 198.84 \\
    \bottomrule
    \end{tabular*}
    \label{tab:appendix_topk_fig}
\end{table}

\noindent\textbf{Router and expert-map diagnostics.}
The router analysis in Fig.~\ref{fig:router} and the expert-map visualization in Fig.~\ref{fig:expert_map} are diagnostic complements to the ablation tables. The router statistics check whether selected experts follow the relabeled physical-category taxonomy, while the expert maps check whether routed updates are spatially concentrated on physically active regions. These diagnostics are not separate training settings; they explain how the full model uses the mechanisms ablated above.

\subsection{Qualitative evidence.}
\begin{figure*}[ht]
    \centering
    \includegraphics[width=1\linewidth]{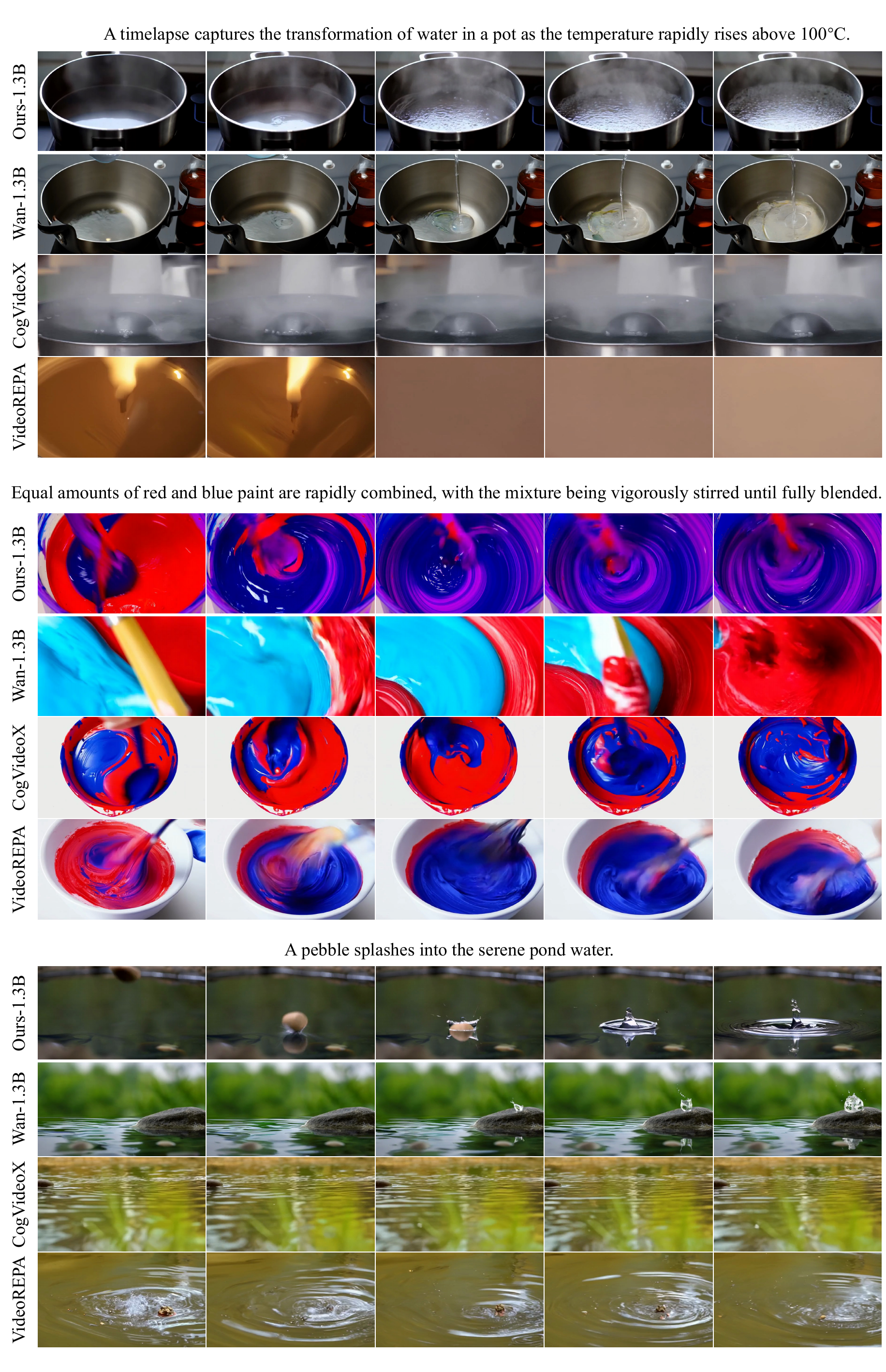}
    \caption{\textbf{Qualitative comparison in the 1.3B--5B setting.} PILA produces more physically plausible motion and interactions under the same prompts.}
    \label{fig:supp_com_new}
\end{figure*}

\begin{figure*}[ht]
    \centering
    \includegraphics[width=1\linewidth]{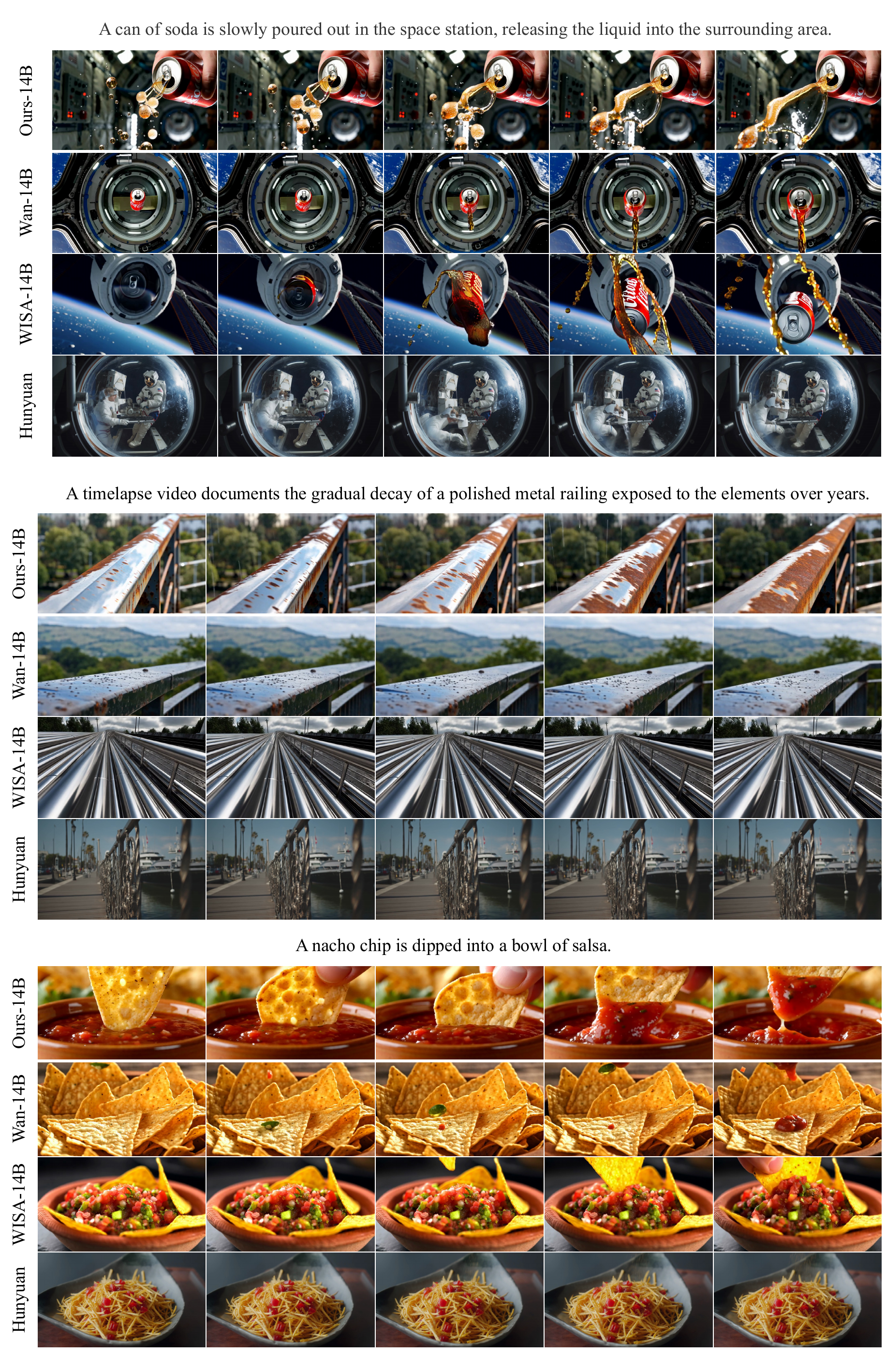}
    \caption{\textbf{Qualitative comparison in the high-capacity setting.} PILA produces more plausible physical evolution under the same prompts.}
    \label{fig:supp_com1}
\end{figure*}

\begin{figure*}[ht]
    \centering
    \includegraphics[width=1\linewidth]{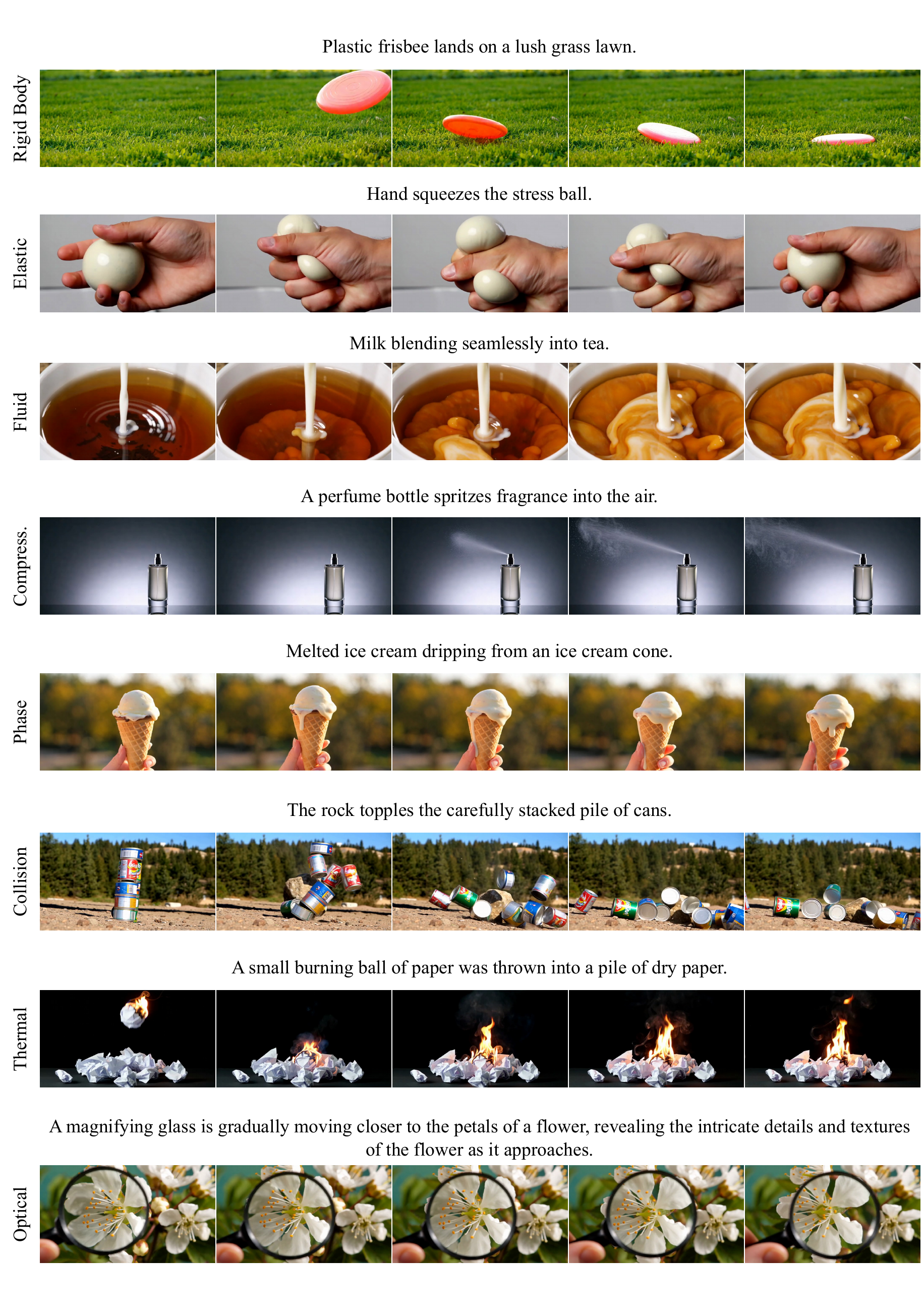}
    \caption{\textbf{Additional PILA generations across physical categories.} Each row shows sampled frames from one physical category in our eight-category taxonomy.}
    \label{fig:supp_show}
\end{figure*}

Figures~\ref{fig:supp_com_new} and~\ref{fig:supp_com1} provide additional qualitative comparisons for the 1.3B--5B and high-capacity settings, respectively. They show that PILA more consistently follows physically plausible object interactions and temporal evolution under the same prompts, complementing the quantitative comparisons in Tables~\ref{tab:quant} and~\ref{tab:quant2}. Figure~\ref{fig:supp_show} further shows category-wise PILA generations across the eight physical categories used by LPMER and CSRC. These examples illustrate the breadth of the generated phenomena, while also motivating the limitations discussed below.

\subsection{Baseline Details}

We compare PILA with the baselines reported in Tables~\ref{tab:quant} and~\ref{tab:quant2}. The comparison is organized by model scale because video-generation quality and physical plausibility are both strongly affected by backbone capacity. All methods are evaluated on the same benchmark prompt sets and with the same metric pipelines described in Sec.~\ref{sec:exp}; no baseline is given access to PILA's operational physical attribute bank or routed residual losses.

\noindent\textbf{Backbone baselines.}

\begin{itemize}[leftmargin=1.2em]
    \item \textbf{Wan 2.1-1.3B}~\cite{wan2025wan}: the frozen text-to-video backbone on which the 1.3B PILA adapter is trained. This row measures the base generator before adding the explicit physical interface, LPMER routing, CSRC residuals, and physics-to-flow correction.
    \item \textbf{Wan 2.2-14B}~\cite{wan2025wan}: the large-scale Wan backbone used as the transfer target in our 14B setting. We include it to separate the effect of model scale from the effect of the transferred PILA adapter.
\end{itemize}

\noindent\textbf{General text-to-video baselines.}

\begin{itemize}[leftmargin=1.2em]
    \item \textbf{CogVideoX-5B}~\cite{yangcogvideox}: a strong open text-to-video diffusion model with an expert-transformer design. It serves as a mid-scale general-generation baseline in the 1.3B--5B comparison group.
    \item \textbf{Hunyuan Video-80G}~\cite{kong2024hunyuanvideo}: a large-scale video generation system included in the high-capacity comparison group. This baseline helps assess whether PILA's physics gains are explained purely by larger model capacity.
\end{itemize}

\noindent\textbf{Physics-aware video-generation baselines.}

\begin{itemize}[leftmargin=1.2em]
    \item \textbf{VideoREPA-2B}~\cite{zhang2025videorepa}: a physics-oriented video generation method that aligns generated dynamics with relational signals from foundation models. We include it as a mid-scale physics-aware baseline.
    \item \textbf{WISA-14B}~\cite{wang2025wisa}: a physics-aware video generation baseline built around world-simulation assistance. It is included in the 14B-scale group and is particularly relevant because our training data is based on WISA-80K, while PILA uses the data through a relabeled eight-category taxonomy rather than through WISA's model architecture.
\end{itemize}

\noindent\textbf{PILA variants.}
\textbf{Ours-1.3B} denotes PILA trained on the Wan 2.1-1.3B backbone with the full physical interface and routed residual objective. \textbf{Ours-14B$^\dagger$} denotes the transfer setting: the adapter trained on Wan 2.1-1.3B is loaded onto the frozen Wan 2.2-14B backbone without Wan 2.2-specific adapter optimization. The dagger is used throughout the tables to mark this no-retraining transfer protocol.

\section{Limitations and Discussions}\label{sec:limit}

\noindent\textbf{Limitations.}
PILA improves physical plausibility by aligning a pretrained flow-matching generator with an operational physical attribute bank, but the bank is designed as a latent interface rather than a full simulator state. Its field slots provide physics-aware proxies shaped by observable motion anchors and operational residuals, which is appropriate for video generation but does not aim to recover calibrated material parameters or exact physical measurements. The category-specific residual constraints also use compact approximations of physical relations, leaving detailed boundary conditions, source terms, and 3D geometry to future extensions. 

\noindent\textbf{Discussion and future work.}
Future work can build on PILA in two directions. First, richer cues such as depth, object tracks, 3D correspondence, or simulator-assisted annotations could further calibrate the physical attribute bank. Second, learned or parameterized physical operators could extend the residual families to more scene-specific boundary conditions and material properties.

